\documentclass[conference]{IEEEtran}
\IEEEoverridecommandlockouts

\usepackage{cite}
\usepackage{amsmath,amssymb,amsfonts}
\usepackage{algorithmic}
\usepackage{graphicx}
\usepackage{textcomp}
\usepackage{xcolor,xspace}
\usepackage{graphicx,multirow,multicol,bm,comment,hyperref,booktabs}

\newcommand*{\eg}{e.g.,\xspace}
\newcommand*{\ie}{i.e.,\xspace}

\newcommand*{\TabletoImage}{\texttt{Table2Image}\xspace}

\newif\ifarxiv
\arxivtrue

\ifarxiv
  \newcommand{\extref}[1]{#1 in the Appendix}
\else
  \newcommand{\extref}[1]{extended version~\cite{t2i_extended}}
\fi

\def\BibTeX{{\rm B\kern-.05em{\sc i\kern-.025em b}\kern-.08em
    T\kern-.1667em\lower.7ex\hbox{E}\kern-.125emX}}

\begin{document}

\title{Table2Image: Lightweight Tabular Learning with Generated Proxy Representations and Reliability Diagnostics}

\author{
\IEEEauthorblockN{
Seungeun Lee\textsuperscript{1},
Kihwan Lee\textsuperscript{2},
Subin Bae\textsuperscript{2},
Sangjun Lee\textsuperscript{2},
Seulbin Lee\textsuperscript{2}, \\
Julia Stoyanovich\textsuperscript{1,*},
Il-Youp Kwak\textsuperscript{3,*},
Seungsang Oh\textsuperscript{2,*}
}

\IEEEauthorblockA{
\textsuperscript{1}New York University, New York, USA \qquad
\textsuperscript{2}Korea University, Seoul, Korea \qquad
\textsuperscript{3}Chung-Ang University, Seoul, Korea \\
\{seungeun.lee, stoyanovich\}@nyu.edu \\
\{lee9843, 2017160022, jimmy1903, slbini01, seungsang\}@korea.ac.kr \\
ikwak2@cau.ac.kr
}

\thanks{\textsuperscript{*}Corresponding authors. Seungsang Oh is the primary corresponding author.}
}

\maketitle

\begin{abstract}

Deep tabular models should ideally balance predictive performance, parameter efficiency, and robustness to imperfect learning signals---properties that are rarely considered jointly. We present Table2Image, a lightweight tabular learning model built around a learned generation pathway that maps tabular inputs into intermediate, structured proxy representations. We additionally examine a variant with variance inflation factor (VIF)-informed initialization, which downweights highly collinear features at the start of training. Across datasets from OpenML-CC18 and TabZilla, Table2Image achieves competitive clean predictive performance while remaining compact relative to several large-scale neural baselines. We further introduce a unified, severity-controlled evaluation protocol under three imperfect learning conditions---irrelevant inputs, corrupted supervision, and unstable shortcut associations---that combines performance-based robustness measures with realization-level instability diagnostics for reliability characterization. Table2Image maintains a favorable balance of performance, robustness, and compactness. Controlled ablations indicate that the learned generation pathway is a key driver of the observed gains.

\end{abstract}


\section{Introduction}
\label{sec:intro}

Tabular data underlies the majority of real-world machine learning applications in science and industry~\cite{shwartz2022tabular,chui2018notes}, and recent deep learning approaches have grown increasingly competitive on tabular tasks~\cite{gorishniy2021revisiting, arik2021tabnet, chen2023trompt, hollmanntabpfn, feuer2024tunetables, gorishniy2024tabm}. These gains, however, typically come at the cost of architectural complexity, large parameter counts, or substantial pretraining and inference overhead. Consequently, gradient-boosted decision trees such as XGBoost~\cite{chen2016xgboost} and LightGBM~\cite{ke2017lightgbm} remain the default choice in practice, offering competitive performance at low computational cost and ease of deployment~\cite{shwartz2022tabular}. Deep learning nonetheless retains a distinct advantage: it supports end-to-end gradient-based optimization and allows tabular inputs to be integrated with unstructured modalities (\eg image or audio inputs) within unified multimodal systems~\cite{gorishniy2021revisiting}. This discrepancy motivates deep tabular models that keep the flexibility of deep learning while staying lightweight and competitive in predictive performance.

Beyond clean predictive performance and model compactness, practical reliability entails evaluating whether models remain effective even under imperfect learning conditions, including irrelevant features, noisy labels, and spurious associations that can bias learning and impair generalization~\cite{liu2022picket,song2022learning,steinmann2024navigating}. 
These failure modes have each received growing attention in isolation, while joint evaluation that compares corruption types, severity levels, and clean performance under one protocol remains limited~\cite{cherepanova2023performance,simonetto2024tabularbench,cheng2025tabfsbench}. 
Existing evaluations characterize robustness typically through performance degradation, leaving open whether the model behavior is actually stable across repeated realizations of the same perturbation---a distinction that performance alone cannot reveal. Together, these motivate a lightweight tabular model examined not only on clean settings but also across diverse types and degrees of imperfect learning.

\textbf{Our contributions:} 
\textbf{Model.} We propose \TabletoImage, which introduces a learned generative transformation between tabular feature processing and classification: it produces a structured auxiliary representation that is subsequently classified by a compact convolutional neural network (CNN). A variance inflation factor (VIF)-informed initialization variant further gives the model a collinearity-aware starting point. It reduces the initial influence of highly correlated features and leaves subsequent end-to-end optimization unconstrained. 
\textbf{Protocol.} We design a severity-controlled protocol under irrelevant features, noisy labels, and shortcut decorrelation, in which within-scenario stability is assessed jointly with standard performance degradation. 
Irrelevant features evaluate sensitivity to non-informative inputs; noisy labels measure the ability to preserve generalization under partially corrupted supervision; and shortcut decorrelation assesses whether a model relies on spurious predictive cues that disappear at test time.
\textbf{Analysis.} Evaluated on OpenML-CC18 and TabZilla, \TabletoImage uses fewer parameters than large-scale neural baselines yet remains competitive with them on both clean and corrupted data. Targeted ablations further clarify the source of the observed gains: removing the learned generation pathway leads to the largest degradation, indicating that the generation pathway is the most consequential component of the design.
Our proposed instability measures show that strong performance under corruption does not necessarily imply stable predictions across perturbations, underscoring the need to evaluate robustness beyond aggregate performance alone. Overall, these results position \TabletoImage as a parameter-efficient tabular model with favorable predictive performance and a broader characterization across complementary reliability dimensions. Our code is available at: \url{https://github.com/duneag2/table2image}.

\section{Related work}
\label{sec:background}

\subsection{Structuring Latent Representations in Tabular Learning}
\label{sec:background:structuring_latent}

Beyond the downstream task loss, auxiliary objectives or structured intermediate representations are often used to shape more useful latent representations in tabular learning. VIME~\cite{yoon2020vime} and SubTab~\cite{ucar2021subtab} rely on reconstruction-based objectives---the former via mask prediction, the latter via subset-level embedding aggregation---to capture feature dependencies. SCARF~\cite{bahriscarf} aligns original and corrupted views to learn perturbation-invariant embeddings. SwitchTab~\cite{wu2024switchtab} decomposes representations into shared and sample-specific salient components, and ReConTab~\cite{chen2023recontab} combines regularized reconstruction with label-aware contrastive learning to suppress redundant features. Collectively, these approaches suggest that auxiliary objectives and structural constraints provide meaningful inductive biases for tabular learning, a principle that motivates the learned generation pathway at the core of \TabletoImage.

\subsection{Tabular-to-Image Representation Learning}

A related line of work converts tabular inputs into image-shaped representations that convolutional architectures then process; Tang {\it et al.}~\cite{tang2023tabular} benchmarks tabular-to-image generation from a visualization-oriented perspective. DeepInsight~\cite{sharma2019deepinsight}, 
SuperTML~\cite{sun2019supertml}, REFINED~\cite{bazgir2019refined}, IGTD~\cite{zhu2021converting}, and TINTO~\cite{castillo2023tinto} each construct two-dimensional layouts from pairwise similarity, correlation, or distance, preserving an explicit correspondence between features and images; the transformed images are then used directly as the classifier input. 
HACNet~\cite{matsuda2024hacnet} moves closer to a learned mapping, training a feature-to-image converter jointly with a CNN. However, it assigns a single representative template per class, limiting representational diversity within each class, and relies on a substantially heavier CNN backbone than ours. \TabletoImage departs from both directions: it learns class-conditioned, instance-varying proxy representations, paired with a lightweight CNN.

\subsection{Evaluating Reliability in Tabular Learning}

Reliability in tabular learning has been studied along several separate axes. One line of work studies robustness to corrupted data and unreliable learning signals: Picket~\cite{liu2022picket} protects tabular pipelines by filtering corrupted instances, and Song {\it et al.}~\cite{song2022learning} studies representation learning to reduce memorization under label noise.
Cherepanova {\it et al.}~\cite{cherepanova2023performance} specifically studies feature selection under extraneous and corrupted features, while Steinmann {\it et al.}~\cite{steinmann2024navigating} surveys how shortcut learning and spurious correlations impair generalization, but do not provide a tabular-specific evaluation.

Another line of work studies distribution and feature shifts: TableShift~\cite{gardner2023benchmarking} benchmarks tabular models under naturally occurring shifts between training and deployment populations, while WhyShift~\cite{liu2023need} studies heterogeneous distribution shifts in tabular data and examines the relationship between in-distribution and shifted performance. TabFSBench~\cite{cheng2025tabfsbench} further compares tabular models across several feature-shift settings and relates degradation to shifted feature importance.
For adversarial robustness, TabularBench~\cite{simonetto2024tabularbench} evaluates model--defense combinations under domain-constrained adversarial attacks.
TabAttackBench~\cite{he2025tabattackbench} benchmarks adversarial attacks for effectiveness and imperceptibility, rather than evaluating the robustness of model architectures or defenses.

Across this body of work, three limitations motivate our evaluation design. \textbf{Fragmented evaluation.} Existing studies generally isolate a single failure mode (\eg feature selection~\cite{cherepanova2023performance} and shortcut learning~\cite{hermann2024foundations}). Differences in datasets, perturbation settings, and evaluation protocols therefore make cross-study comparisons of reliability difficult. 
\textbf{Performance-centered assessment.} Robustness is typically characterized through predictive performance under corruption, with less attention to how consistently a model behaves across realizations of the same perturbation setting; parameter efficiency is also rarely considered jointly with robustness.
\textbf{Limited architectural insight.} Prior interventions are largely scenario-specific---feature selectors, corruption detectors, augmentations, or attack--defense pairs---offering limited evidence about which underlying architectures or representation learning mechanisms are reliable across multiple forms of imperfect learning. We therefore evaluate \TabletoImage and representative baselines under a common severity-controlled protocol that combines clean predictive performance, parameter count, performance under corruption, and realization-level instability.

\begin{figure}[!ht]
\begin{center}
\centerline{\includegraphics[width=\linewidth]{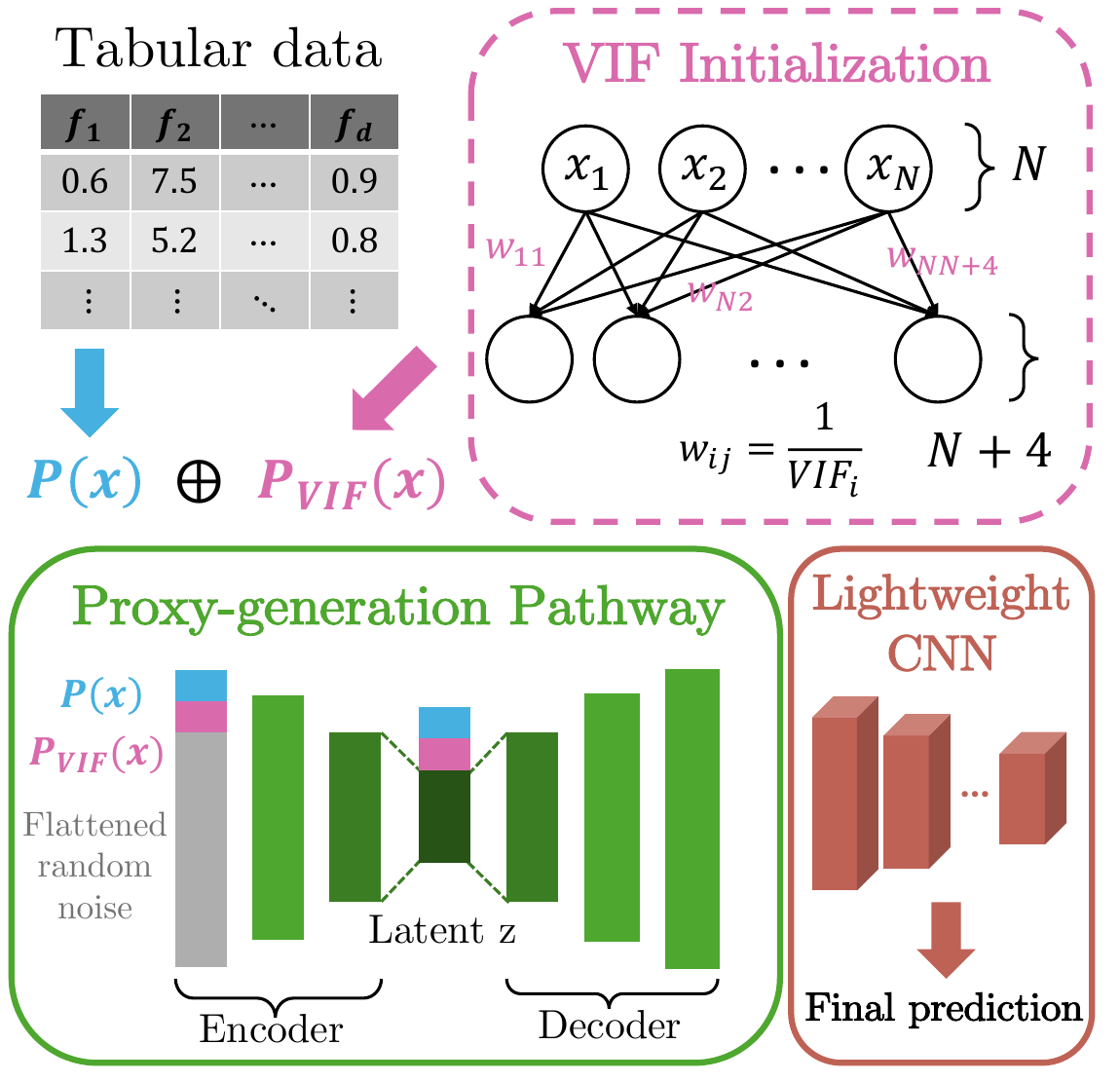}} 
\caption{Overview of the \TabletoImage framework: a tabular feature processor, learned proxy-generation pathway, and lightweight CNN, with the optional VIF-informed initialization shown separately.}
\label{fig1}
\end{center}
\end{figure}

\section{Methodology}
\label{sec:models}

We present the \TabletoImage framework and its VIF-informed variant, then introduce the robustness scenarios and evaluation metrics underlying our reliability diagnostics.

\subsection{\TabletoImage Framework}
\label{sec:framework}

The framework consists of three components: a tabular feature processor, a proxy representation generator, and a CNN classifier (Figure~\ref{fig1}). We consider two strategies for generating proxy targets: one uses externally provided image instances and the other constructs class-aware proxy targets from the data itself. 

\textbf{External image-based proxy targets.} 
Proxy targets are sampled from an external image set $I$ based on the number of classes $C$ in the tabular dataset $X$. We use FashionMNIST~\cite{xiao2017fashion} for $C \leq 10$, and combine FashionMNIST with MNIST~\cite{6296535} to support up to 20 classes for $C > 10$ ($C > 20$ are left for future work).
Each instance $x \in X$ with label $y \in \{0, \dots, C-1\}$ is paired with an image $i_x \in I$ drawn at random from the same class. This randomized, one-to-many mapping exposes the model to diverse input--output pairs rather than a single fixed prototype per class---a design choice we validate empirically in Section~\ref{sec:results:single}.

\textbf{Internally constructed proxy targets.} The internal mode takes a different view; rather than conveying image semantics, it builds a structured representation that combines a class-level template with instance-specific variation. For each sample $(x,y)$, we define $q(x,y)=(1-\alpha)T_y+\alpha V(x),$ where $T_y$ is a template shared across all samples in class $y$, $V(x)$ is an instance-specific feature map constructed directly from the tabular input $x$, and $\alpha\in[0,1]$ controls their relative contribution ($0.35$ by default). This construction is intentionally simple, as its role is to provide a structured auxiliary learning signal rather than encode image semantics. The label $y$ is used only to construct the proxy target during training and is not provided to the generator at inference. This mode imposes no restriction on the number of classes, and the resulting $\mathbb{R}^{28\times28}$ layout is a convenient instantiation compatible with the downstream CNN (see Figure~\ref{fig:internal}).

The class template $T_y\in[0,1]^{28\times28}$ combines a Gaussian component $G_y$, which provides coarse spatial separation between classes, with a periodic component $P_y$ that adds finer class-specific structure. For $(u_{ij},v_{ij})\in[-1,1]^2$ denoting the coordinates of pixel $(i,j)$, we place the Gaussian centers uniformly on a circle of radius $0.45$, with center
$(c_1^{(y)}, c_2^{(y)})=(0.45\cos(2\pi y/C),\,0.45\sin(2\pi y/C))$ for class $y$:
$G_y(i,j)=\exp(-((u_{ij}-c_1^{(y)})^2+(v_{ij}-c_2^{(y)})^2)/0.18).$
These centers may overlap as $C$ increases; we therefore complement $G_y$ with
$
P_y(i,j)
=
\tfrac{1}{2}(\sin((y+1)\pi\psi_y(u_{ij},v_{ij}))+1),
$
where 

{\small
$$
\psi_y(u,v)
=
\begin{cases}
u, & y\bmod 4=0,\\
v, & y\bmod 4=1,\\
u+v, & y\bmod 4=2,\\
\sqrt{u^2+v^2}, & y\bmod 4=3,
\end{cases}
$$
}

producing horizontal, vertical, diagonal, and radial patterns, respectively. The frequency factor $(y+1)\pi$ further gives each class a distinguishing spatial density, with larger values producing denser patterns. The final class template is
$T_y=\operatorname{Norm}(0.55G_y+0.45P_y),$ where $\operatorname{Norm}(\cdot)$ denotes min--max normalization to $[0,1]$.

Variation within a class comes from $V(x)$, whose role is simply to introduce small sample-specific deviations around the shared class template. It is built directly from the tabular features themselves: the features are tiled until they fill the $28\times28=784$ entries, passed through $\tanh$, min--max normalization, and smoothed with $3\times3$ average pooling.

\textbf{Tabular feature processing.}
Each input $x\in\mathbb{R}^{N}$ passes through a lightweight MLP, $\mathtt{P}(x) = \mathtt{R(FC_2(R(FC_1}(x)))), $
where $\mathtt{FC_1}$ is a fully connected layer that expands the dimensionality to $N\!+\!4$, $\mathtt{FC_2}$ reduces it back to $N$, and $\mathtt{R}$ denotes the ReLU activation function. This transformation gives the generator a learned tabular representation before mapping it into the structured proxy space.

\textbf{Proxy representation generation.}
For each instance, we independently sample a random input $r \sim \mathcal{U}(0,1)^{28\times28}$ on every forward pass and combine it with the tabular embedding to form the proxy representation. 

The encoder concatenates $\mathtt{P}(x)$ with the flattened random noise vector $\mathtt{flatten}(r) \in \mathbb{R}^{784}$ and maps them into a latent representation $z \in \mathbb{R}^N$:
$$z=\mathtt{R}\!\left(\mathtt{FC}_{4}\!\left(\mathtt{R}\!\left(\mathtt{FC}_{3}\bigl(\mathtt{flatten}(r) \oplus \mathtt{P}(x)\bigr)\right)\right)\right),$$
where $\oplus$ denotes concatenation. 
The generator $\mathtt{G}$ then combines $z$ with $\mathtt{P}(x)$ to produce the proxy representation, reshaping the $784$-dimensional output into $28 \times 28$: 
$$
\mathtt{G}(\mathtt{P}(x), r) = \mathtt{reshape}\!\left( \sigma\!\left(\mathtt{FC}_{6}\!\left(\mathtt{R}\!\left(\mathtt{FC}_{5}(z \oplus \mathtt{P}(x))\right)\right)\right)\right) \in \mathbb{R}^{28 \times 28},
$$
with $\mathtt{FC}_{3}$ and $\mathtt{FC}_{5}$ both using a 128-dimensional hidden representation and $\sigma$ denoting sigmoid.
The latent representation $z$ is designed to be compact yet class-informative enough that the resulting representations exhibit class-specific structure while varying across instances (Section~\ref{sec:experiments:vis}).

\textbf{CNN for classification.}
A lightweight CNN classifies the generated representation $\mathtt{G}(\mathtt{P}(x),r)$: two convolutional layers with ReLU activations and max pooling, followed by two fully connected layers (full architectural details are in our code):
\begin{gather*}
z'=\mathtt{Max_2(R(Conv_2(Max_1(R(Conv_1(}\mathtt{G}(\mathtt{P}(x),r))))))) \\
\mathtt{CNN}(\mathtt{G}(\mathtt{P}(x),r))=(\mathtt{FC_8(Drop(R(FC_7(flatten}(z')))))) 
\end{gather*}

\textbf{Training objective.}
We train the model end-to-end with the AdamW
optimizer~\cite{loshchilov2018decoupled}, minimizing
$
\mathcal{L}_{\mathrm{total}}
=
\mathcal{L}_{\mathrm{cls}}
+
\lambda_{\mathrm{recon}}\mathcal{L}_{\mathrm{recon}}.
$
The classification term is $
\mathcal{L}_{\mathrm{cls}}
=
\operatorname{CE}(\mathtt{CNN}(\mathtt{G}(\mathtt{P}(x),r)),y),$ the standard cross-entropy loss, while the proxy reconstruction term
$
\mathcal{L}_{\mathrm{recon}}
=
\left\lVert
\mathtt{G}(\mathtt{P}(x),r)-p_x^{*}
\right\rVert_2^2,
$
anchors the generated representation to its proxy target $p_x^{*}$---the paired image $i_x$ for externally constructed targets, or $q(x,y)$ for internally constructed ones.

\subsection{Table2Image-VIF: VIF Initialization} \label{sec:method:vif}

\subsubsection{Multicollinearity and variance inflation factor (VIF)}

Multicollinearity arises when two or more independent variables in a dataset are linearly related, with imperfect multicollinearity being a common case in real-world datasets. While there is no consensus on whether collinear variables should simply be removed~\cite{gujarati2003multicollinearity}, the variance inflation factor (VIF) offers a way to quantify the degree of multicollinearity.
The VIF of the $i$-th input feature variable $X_i$ is defined as
$\mathtt{VIF}_i = 1/(1-R_i^2)$
where $R_i^2$ is the coefficient of determination from regression of $X_i$ on the other remaining features~\cite{james2013introduction}; higher values indicate greater collinearity, with $\mathtt{VIF}_i > 10$ generally considered problematic.
In image or text domains, feature correlations often carry semantic meaning and, in deep learning, multicollinearity can be implicitly addressed by dropout~\cite{srivastava2014dropout} or regularization techniques. 
In contrast, tabular data, characterized by high dimensionality and sparse feature distributions, may face severe multicollinearity, which in turn can destabilize models. 

\subsubsection{Table2Image-VIF} 

\emph{Table2Image-VIF} extends \TabletoImage with a VIF-informed transformation $\mathtt{P_{VIF}}(x)$, added alongside the original $\mathtt{P}(x)$. The downstream architecture is unchanged; the only addition is how the first MLP layer is initialized:
$w_{ij} = 1/\mathtt{VIF}_i,$
where {\small $1 \leq i \leq N,\, 1 \leq j\leq N+4,$} so that features with higher collinearity start out with smaller weights. From there, optimization proceeds as usual; VIF values shape only where the training starts, not where it can end up.
The final representation in \emph{Table2Image-VIF} concatenates the original $\mathtt{P}(x)$ with this VIF-initialized embedding:
$\mathtt{P}(x) \oplus \mathtt{P_{VIF}}(x). $
This keeps the original representation intact while complementing it with a collinearity-aware view of the same features (see Section~\ref{sec:experiments:vif} for an empirical validation).

\subsection{Robustness Scenarios}
\label{sec:robustness_scenarios}

We consider three robustness scenarios capturing complementary sources of imperfect learning signals: irrelevant features, noisy labels, and shortcut decorrelation. 
The first two have been previously considered in tabular learning, but their behavior across systematically varied severity and model families remains insufficiently studied. Shortcut learning has mainly been examined in other domains; we adapt it to tabular data through a controlled train--test decorrelation protocol. Together, these scenarios encompass input-side distraction, supervision corruption, and train--test association shift.

\textbf{Irrelevant features (IF).} Following prior work on random features~\cite{cherepanova2023performance}, we append Gaussian random features to the original inputs $X_{\{\mathrm{train}, \mathrm{test}\}} \in \mathbb{R}^{n_{\{\mathrm{train}, \mathrm{test}\}} \times N}
$, leaving all original features unchanged. At irrelevant feature ratio $\rho_\text{r} \in \{0.05, 0.10, 0.15, 0.20\}$, we sample $R_{\{\mathrm{train,test}\}}$ i.i.d.\ with $\operatorname{round}(\rho_{\text{r}}N)$ independent features from $\mathcal{N}(0,1)$, and concatenate them with the original inputs:
$\left[
X_{\{\mathrm{train,test}\}}
\;\middle\|\;
R_{\{\mathrm{train,test}\}}
\right],
$ where $\|$ denotes column-wise concatenation.
$R_{\mathrm{train}}$ and $R_{\mathrm{test}}$ are drawn independently, so noise patterns learned during training carry no signal at test time; this isolates whether a model preserves predictive performance as the number of task-irrelevant input dimensions increases.

\textbf{Noisy labels (NL).} Following~\cite{zhang2017understanding,arpit2017closer,algan2020label}, we corrupt only the training labels 
$
y_{\mathrm{train}}
\in
\{0,\ldots,C-1\}^{n_{\mathrm{train}}},
$ leaving $X_{\mathrm{train}}$, $X_{\mathrm{test}}$, and clean test labels unchanged. For each training instance, we independently sample
$
m_i
\sim
\operatorname{Bernoulli}(\rho_\text{n})
$
at the label noise ratio (\ie the expected proportion of training labels replaced by incorrect class labels):
$
\rho_\text{n}
\in
\{0.05,0.10,0.15,0.20\},
$
with $\mathbb{E}\left[\sum_{i=1}^{n_{\mathrm{train}}} m_i\right] =n_{\mathrm{train}}\rho_\text{n}.$ 
The corrupted label is defined as:

{\small
$$
\widetilde{y}_i
=
\begin{cases}
y_i, & m_i=0,\\
(y_i+\delta_i)\bmod C, & m_i=1,
\end{cases}
\; \delta_i
\sim
\operatorname{Uniform}\{1,\ldots,C-1\}
$$
}

so that selected labels are reassigned to a different class, ensuring $\widetilde{y}_i\neq y_i$. Models train on $(X_{\mathrm{train}}, \widetilde{y}_{\mathrm{train}})$ but are evaluated on the clean test set, measuring how well generalization survives increasingly corrupted supervision. 

\textbf{Shortcut decorrelation (SD).} Following~\cite{geirhos2020shortcut} and~\cite{hermann2024foundations}, we add one synthetic shortcut feature that is predictive of the target during training but uninformative at test time. This scenario is governed by two parameters: its training-time predictivity $\rho_\text{s} \in \{0.60,0.70,0.80,0.90,0.95,1.00\},$ and availability $a_\text{s} \in \{0.5,1.0,2.0\}$ (adapted from~\cite{hermann2024foundations}). For example, $\rho_\text{s}=0.60$ and $a_\text{s}=2.0$ produce a shortcut that agrees with the target label for $60\%$ of the training instances and whose normalized value is multiplied by $2.0$.
For each training instance, $m_i \sim \operatorname{Bernoulli}(\rho_\text{s})$ gives

{\small
$$
h_i
=
\begin{cases}
y_i, & m_i=1,\\
(y_i+\delta_i)\bmod C, & m_i=0,
\end{cases}
\; \delta_i \sim \operatorname{Uniform}\{1,\ldots,C-1\}
$$
}

used only to construct the shortcut feature $s_i^{\mathrm{train}} = a_\text{s}h_i/(C-1)$, not to replace the true label. At test time, $s_i^{\mathrm{test}} = a_\text{s} h_i^{\mathrm{test}}/(C-1)$ is instead built from independently sampled $h_i^{\mathrm{test}} \sim \operatorname{Uniform}\{0,\ldots,C-1\}$, thereby removing the shortcut--label association while preserving its scale. The augmented inputs are $\left[
X_{\{\mathrm{train, test}\}}
\;\middle\|\;
s_{\{\mathrm{train, test}\}}\right]$.

\subsection{Evaluation Metrics}
\label{sec:evaluation_metrics}

We evaluate \TabletoImage and comparative models along three complementary perspectives: clean predictive performance, scenario-agnostic performance-based robustness, and scenario-specific realization-level instability diagnostics. The first two assess predictive performance on the original data and its degradation across perturbation severity, respectively (Section~\ref{sec:experiments:eval_metrics}). The third---our proposed instability diagnostics---quantifies within-scenario behavior variability across repeated perturbation realizations at a fixed severity level. 

For two independently sampled realizations $w$ and $\ell$ of the same perturbation setting, let
$p_i^{(w)},p_i^{(\ell)}\in[0,1]^C$
denote the predicted class probabilities for test instance $i$, with $\widehat{y}_i^{(k)}=\arg\max_c p_{i,c}^{(k)}$ for $k\in\{w,\ell\}$.
We define the pairwise instability

{\small
$$
d^{(w,\ell)}(i)
=
\begin{cases}
1,
& \widehat{y}_i^{(w)} \neq \widehat{y}_i^{(\ell)}, \\
\frac{1}{2}
\left\lVert
p_i^{(w)}-p_i^{(\ell)}
\right\rVert_1,
& \widehat{y}_i^{(w)} = \widehat{y}_i^{(\ell)}.
\end{cases}
$$
}

A decision change counts as maximal instability (score $1$); otherwise,
instability reflects the discrepancy between the two predictive distributions.
For each severity setting, we average $d^{(w,\ell)}(i)$ over all test instances
and realization pairs:

{\small
$$
\frac{1}{n_\text{test}}
\sum_{i=1}^{n_\text{test}}
\frac{2}{K(K-1)}
\sum_{1\leq w<\ell\leq K}
d^{(w,\ell)}(i).
$$
}

We then summarize each instability metric over its evaluated severity range by normalized integration, using the corresponding severity dimensions for each scenario, and refer to this quantity as \emph{Decision-aware Prediction Instability} ($\operatorname{DPI}$)
for irrelevant features, \emph{Label Corruption Instability} ($\operatorname{LCI}$) for noisy labels, and \emph{Shortcut Assignment Instability} ($\operatorname{SAI}$) for shortcut decorrelation. All values lie in $\left[0,1\right]$ with lower values indicating greater stability. Instability is computed over all $K$ realizations separately within each training seed and then averaged across the three training seeds; this isolates realization-level variability from training stochasticity.

\section{Experimental Evaluation}
\label{sec:experiments}

\subsection{Experimental Setup}
For the external proxy targets, which support classification up to 20 target classes, we evaluate \TabletoImage (\texttt{T2I}) on 67 datasets from OpenML-CC18~\cite{bischl2openml} and 34 hard datasets from the TabZilla benchmark suite~\cite{mcelfresh2024neural}; five OpenML-CC18 datasets and two TabZilla datasets exceeding the 20-class limit are excluded. We compare against a broad set of baselines: Logistic Regression~\cite{cox1958regression}, SVM~\cite{cortes1995support}, Random Forest~\cite{liaw2002classification}, three GBDT models--XGBoost~\cite{chen2016xgboost}, LightGBM~\cite{ke2017lightgbm}, and CatBoost~\cite{prokhorenkova2018catboost}--and six neural models: MLP (three layers; not parameter-matched), MLP-PLR~\cite{gorishniy2022embeddings}, FT-Transformer~\cite{gorishniy2021revisiting}, TabPFN (v1)~\cite{hollmanntabpfn}, TuneTables~\cite{feuer2024tunetables}, and TabM~\cite{gorishniy2024tabm}.

To isolate robustness arising from the model architecture itself rather than from externally sourced, class-conditioned supervision, all robustness experiments instead use internally constructed proxy targets.
For clean evaluation, we compare against Logistic Regression, SVM, XGBoost, CatBoost, a parameter-matched MLP (three layers; matched to \texttt{T2I}), FT-Transformer, and TabTransformer, using the 16 datasets shared between OpenML-CC18 and TabZilla. For the robustness scenarios, we select representative models from each family: SVM, XGBoost (GBDTs), the parameter-matched MLP (neural networks), and TabTransformer~\cite{huang2020tabtransformer} (Transformers). TabTransformer is included for its demonstrated robustness to noisy features. The complete dataset list, OpenML identifiers, preprocessing procedures, and detailed configurations (\eg hyperparameter search ranges) are provided in our code. We use an 8:2 train-test split, with the results averaged over three runs. Neural models are trained with a batch size of 64 for 100 epochs. Experiments are conducted on an Intel Xeon Platinum 8592+ CPU (128 cores, 512\,GB RAM) and an NVIDIA A100 GPU (80\,GB VRAM).

\begin{table*}[!ht]
\centering
\caption{Performance comparison of \TabletoImage (\texttt{T2I}) and Table2Image-VIF (T2I-VIF) with 12 high-performing models under external image-based proxy targets on (a) OpenML-CC18 and (b) TabZilla. TabPFN is evaluated only on applicable datasets due to its computational limitations.  Win counts include ties.}
\label{tab:performance_1}
\renewcommand{\arraystretch}{1.15}
\setlength{\tabcolsep}{4.5pt}

\begin{minipage}[t]{0.495\textwidth}
\centering
\textbf{(a) OpenML-CC18}

\vspace{0.05in}

\begin{tabular}{lcccc}
\toprule
\multirow{2}{*}{Model}
& \multirow{2}{*}{Avg ACC}
& \multirow{2}{*}{Avg AUC}
& \multicolumn{2}{c}{\# Wins} \\
\cmidrule(lr){4-5}
& & & ACC & AUC \\
\midrule
\textbf{T2I-VIF}
& \textbf{0.879}
& \textbf{0.922}
& \textbf{27}
& \textbf{26} \\
\textbf{{T2I}}
& 0.877
& 0.920
& 23
& 23 \\
XGBoost
& 0.868
& 0.876
& 7
& 3 \\
TuneTables
& 0.865
& 0.915
& 5
& 6 \\
CatBoost
& 0.863
& 0.915
& 5
& 11 \\
LightGBM
& 0.861
& 0.912
& 13
& 14 \\
Random Forest
& 0.853
& 0.911
& 9
& 6 \\
TabM
& 0.841
& 0.896
& 8
& 10 \\
FT-Transformer
& 0.831
& 0.902
& 5
& 11 \\
SVM
& 0.784
& 0.875
& 3
& 3 \\
Logistic Regression
& 0.783
& 0.876
& 2
& 3 \\
MLP-PLR
& 0.781
& 0.782
& 2
& 1 \\
MLP
& 0.673
& 0.704
& 1
& 0 \\
TabPFN (v1)
& --
& --
& 4
& 4 \\
\bottomrule
\end{tabular}
\end{minipage}
\hfill
\begin{minipage}[t]{0.495\textwidth}
\centering
\textbf{(b) TabZilla}

\vspace{0.05in}

\begin{tabular}{lcccc}
\toprule
\multirow{2}{*}{Model}
& \multirow{2}{*}{Avg ACC}
& \multirow{2}{*}{Avg AUC}
& \multicolumn{2}{c}{\# Wins} \\
\cmidrule(lr){4-5}
& & & ACC & AUC \\
\midrule
\textbf{T2I-VIF}
& \textbf{0.862}
& 0.905
& 8
& 7 \\
\textbf{{T2I}}
& 0.857
& \textbf{0.906}
& \textbf{9}
& 3 \\
TuneTables
& 0.854
& 0.894
& 1
& 3 \\
LightGBM
& 0.845
& 0.899
& 7
& \textbf{8} \\
CatBoost
& 0.842
& 0.902
& 5
& 5 \\
Random Forest
& 0.838
& 0.887
& 5
& 5 \\
XGBoost
& 0.837
& 0.902
& 4
& 1 \\
FT-Transformer
& 0.824
& 0.900
& 3
& 5 \\
TabM
& 0.790
& 0.888
& 5
& 2 \\
SVM
& 0.772
& 0.857
& 1
& 0 \\
Logistic Regression
& 0.744
& 0.839
& 1
& 1 \\
MLP-PLR
& 0.736
& 0.752
& 1
& 0 \\
MLP
& 0.625
& 0.641
& 1
& 0 \\
TabPFN (v1)
& --
& --
& 1
& 1 \\
\bottomrule
\end{tabular}
\end{minipage}

\end{table*}

\subsection{Evaluation Metrics}
\label{sec:experiments:eval_metrics}

We evaluate each model along clean, scenario-agnostic, and scenario-specific axes. For clean predictive performance, we report average accuracy (Avg ACC), average area under the ROC curve (Avg AUC), and win count (including ties).
For scenario-agnostic robustness, we compute the \emph{area under the performance--severity curve} ($\operatorname{APS}$), \emph{worst-severity performance} ($\operatorname{WSP}$), and \emph{$\tau$-degradation percentage}. Let $m(\rho)$ denote either ACC or AUC at severity level $\rho$; the normalized $\operatorname{APS}$ is
$
\operatorname{APS}_{m}
=
\int_{0}^{\rho_{\max}} m(\rho) /  {\rho_{\max}}\,d\rho,
$
with higher values indicating better performance across perturbation severities. This metric summarizes absolute predictive performance across the evaluated severity range and therefore reflects both the clean starting point ($\rho=0$) and subsequent performance under perturbation. \emph{Worst-severity performance} ($\mathrm{WSP}$) is the ACC and AUC at the maximum evaluated severity $\rho_{\max}$, set to $\rho_\text{r}=0.20$ for irrelevant features, $\rho_\text{n}=0.20$ for noisy labels, and $(\rho_\text{s},a_\text{s})=(1.0,2.0)$ for shortcut decorrelation.

To complement $\text{APS}$ and $\text{WSP}$ with a relative degradation view, we define the \emph{$\tau$-degradation percentage}, which captures how much perturbation a model tolerates before a given performance drop: 
$
\rho_{\tau}^{(m)}
=
100 \times
{
\min
\left\{
\rho \in \mathcal{P}
\mid
m(\rho) \leq m_\mathrm{clean}-\tau
\right\}
}/{
\rho_{\max} 
},
$
where $\mathcal{P}$ is the set of evaluated severities, $m_\mathrm{clean}$ is the clean performance. If the drop is never reached within the evaluated range, we set $\rho_{\tau}^{(m)}=100$. Larger values indicate a greater perturbation severity is required to degrade performance by $\tau$, and thus greater robustness. For shortcut decorrelation, performance and instability are averaged over $a_s$ at each $\rho_s$ before being summarized over $\rho_s$, while $\tau$-degradation is reported as the threshold-crossing pair $(\rho_s,a_s)$.
Finally, we use the scenario-specific instability metrics from Section~\ref{sec:evaluation_metrics}---${\operatorname{DPI}}$, ${\operatorname{LCI}}$, and ${\operatorname{SAI}}$---to characterize prediction variability across repeated realizations within each robustness scenario.

\subsection{Experimental Results}
\label{sec:experiments:results}

Our experiments address the following research questions: clean predictive performance and parameter efficiency (\textbf{RQ1}), robustness and stability under imperfect learning signals (\textbf{RQ2}), contributions of design components and compatibility of the generation pathway across encoders (\textbf{RQ3}), and the structure of the learned proxy representations (\textbf{RQ4}).

\subsubsection{Predictive performance and parameter efficiency}
\label{sec:results:clean}

\paragraph{External Proxy Targets}

\texttt{T2I} and T2I-VIF achieve competitive clean performance across both OpenML-CC18 and TabZilla (Table~\ref{tab:performance_1}). On OpenML-CC18, T2I-VIF attains the highest average accuracy and AUC, together with the best average rank across datasets for both metrics; model differences are significant under the Friedman test, excluding TabPFN (ACC: $p<0.001$, Kendall's $W=0.297$; AUC: $p<0.001$, Kendall's $W=0.316$). On TabZilla, T2I-VIF achieves the highest average accuracy with the second-best average rank, while \texttt{T2I} achieves the highest average AUC with the third-best average rank; differences are again significant (ACC: $p<0.001$, Kendall's $W=0.253$; AUC: $p<0.001$, Kendall's $W=0.288$), consistent with the competitive performance of both variants across benchmark suites. See~\extref{Table~\ref{tab:pairwise_stats}} for detailed statistics.

\paragraph{Internal Proxy Targets}

\texttt{T2I} shows favorable clean performance across the compared models (Table~\ref{tab:clean_performance_compact}). For accuracy, model differences are significant under the Friedman test ($p<0.001$, Kendall's $W=0.269$), with \texttt{T2I} achieving the best average rank (2.44). It improves over the parameter-matched MLP by 0.81 percentage points on average and achieves higher accuracy on 10 of 16 datasets, while also showing statistically significant Holm-corrected pairwise differences ($p<0.05$) relative to TabTransformer, XGBoost, and SVM. For AUC, \texttt{T2I} likewise achieves the best average rank (2.88), although differences among the leading models are comparatively small. Overall, these results indicate a modest but favorable clean performance advantage for \texttt{T2I}, while the comparable performance of T2I-VIF suggests that internally constructed proxy targets retain comparable performance.

\begin{table}[!b]
\centering
\caption{Clean test performance averaged across datasets, using internal proxy targets. For win counts, ties are included; T2I-VIF is reported for reference and excluded from the win-count comparison. MLP denotes Parameter-matched MLP.}
\label{tab:clean_performance_compact}
\setlength{\tabcolsep}{3pt}
\begin{tabular}{lcccc}
\toprule
\multirow{2}{*}{Model} & \multirow{2}{*}{Avg ACC} & \multirow{2}{*}{Avg AUC} & \multicolumn{2}{c}{\# Wins} \\
\cmidrule(lr){4-5}
 & & & ACC & AUC \\
\midrule
{T2I} & $\mathbf{0.876}$ & $\mathbf{0.936}$ & $\mathbf{6}$ & $\mathbf{4}$ \\
T2I-VIF & $0.874$ & $0.935$ & $-$ & $-$ \\
MLP & $0.868$ & $0.934$ & $3$ & $2$ \\
FT-Transformer & $0.865$ & $0.934$ & $1$ & $1$ \\
CatBoost & $0.845$ & $0.932$ & $2$ & $\mathbf{4}$ \\
XGBoost & $0.843$ & $0.928$ & $0$ & $\mathbf{4}$ \\
TabTransformer & $0.837$ & $0.922$ & $4$ & $1$ \\
SVM & $0.833$ & $0.909$ & $1$ & $1$ \\
LR & $0.801$ & $0.891$ & $0$ & $0$ \\
\bottomrule
\end{tabular}
\end{table}

\paragraph{Parameter-performance trade-off}

\begin{figure}[!ht]
\begin{center}
\centerline{\includegraphics[width=\linewidth]{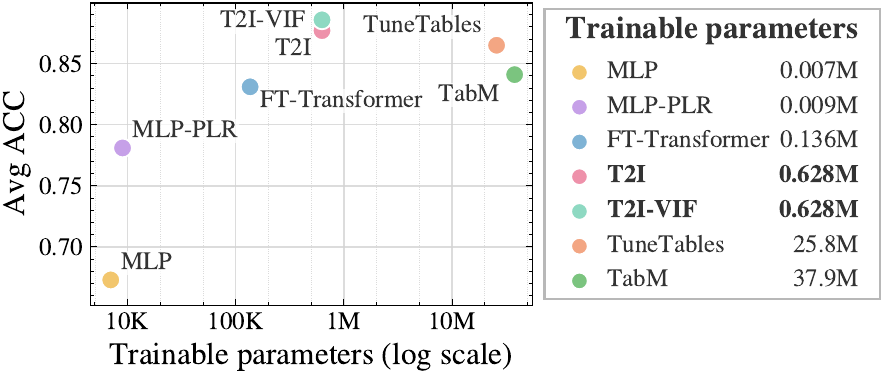}} 
\caption{Average ACC on OpenML-CC18 datasets versus the number of trainable parameters, with parameter counts shown on a logarithmic scale. The corresponding trainable parameter counts, with two classes and a tabular input dimension of $N=78$, are provided separately.}
\label{fig:x_param_y_perf_openml_legend}
\end{center}
\end{figure}

\texttt{T2I} achieves competitive performance with a compact parameter budget: 0.628M trainable parameters (Figure~\ref{fig:x_param_y_perf_openml_legend}), compared with TuneTables (25.8M) and TabM (37.9M), although more than MLP, MLP-PLR, and FT-Transformer, suggesting a favorable balance between model capacity and predictive performance.

\paragraph{Comparison with existing tabular-to-image models} 
We compare \texttt{T2I} (external proxy targets) against HACNet~\cite{matsuda2024hacnet}, DeepInsight~\cite{sharma2019deepinsight}, and IGTD~\cite{zhu2021converting}; since IGTD only produces converted image representations, we pair it with the CNN used in \texttt{T2I}. On the full OpenML-CC18 datasets, HACNet reaches an average accuracy of 0.800, whereas \texttt{T2I} records 0.877. DeepInsight and IGTD are each evaluated on the subset of OpenML-CC18 meeting their respective minimum sample size requirements; \texttt{T2I} reaches an average accuracy of 0.888 vs.\ 0.803 of DeepInsight, and on the IGTD-compatible subset, 0.995 (\texttt{T2I}) vs.\ 0.971 (IGTD).

\subsubsection{Robustness and Stability under imperfect learning}
\label{sec:results:robustness}

\paragraph{Robustness across severity}

\ifarxiv
  
\begin{table}[!hb]
\centering
\caption{Robustness results across three corruption scenarios. We compute the metrics for each dataset and run and report the average. Higher values are better for all metrics, and MLP denotes parameter-matched MLP. 
}
\label{tab:robustness_summary}
\setlength{\tabcolsep}{5pt}
\renewcommand{\arraystretch}{1.10}

\begin{tabular}{@{}llcc@{}}
\toprule
Scenario & Model
& $\operatorname{APS}_{\mathrm{ACC/AUC}}\uparrow$
& $\operatorname{WSP}_{\mathrm{ACC/AUC}}\uparrow$ \\
\midrule

\multirow{6}{*}{IF}
& T2I            & $\mathbf{0.863}/\mathbf{0.929}$ & $0.854/0.924$ \\
& T2I-VIF        & $\mathbf{0.863}/\mathbf{0.929}$ & $\mathbf{0.855}/\mathbf{0.925}$ \\
& MLP            & $0.857/0.927$ & $0.851/0.922$ \\
& TabTransformer & $0.842/0.923$ & $0.840/0.922$ \\
& XGBoost        & $0.839/0.924$ & $0.837/0.921$ \\
& SVM            & $0.828/0.906$ & $0.824/0.904$ \\
\midrule

\multirow{6}{*}{NL}
& T2I            & $\mathbf{0.846}/\mathbf{0.922}$ & $\mathbf{0.822}/\mathbf{0.910}$ \\
& T2I-VIF        & $0.844/0.920$ & $0.815/0.908$ \\
& MLP            & $0.844/0.919$ & $\mathbf{0.822}/0.904$ \\
& XGBoost        & $0.828/0.910$ & $0.807/0.890$ \\
& SVM            & $0.826/0.899$ & $0.817/0.891$ \\
& TabTransformer & $0.824/0.911$ & $0.812/0.902$ \\
\midrule

\multirow{6}{*}{SD}
& T2I            & $\mathbf{0.853}/\mathbf{0.920}$ & $0.721/0.812$ \\
& T2I-VIF        & $\mathbf{0.853}/\mathbf{0.920}$ & $\mathbf{0.722}/\mathbf{0.813}$ \\
& MLP            & $0.849/0.919$ & $0.718/0.804$ \\
& TabTransformer & $0.818/0.905$ & $0.646/0.752$ \\
& SVM            & $0.810/0.890$ & $0.672/0.758$ \\
& XGBoost        & $0.797/0.898$ & $0.460/0.701$ \\

\bottomrule
\end{tabular}
\end{table}

\begin{table}[!t]
\centering
\caption{$\rho_{\tau}^{\mathrm{AUC}}$ (\%) values across three corruption scenarios for $\tau \in \{0.03, 0.05, 0.07\}$. Higher values are better; values are reported as normalized $(\rho_s,a_s)$ percentages for SD.}
\label{tab:tau_auc}
\setlength{\tabcolsep}{2.3pt}
\renewcommand{\arraystretch}{1.04}

\begin{tabular}{@{}lcccccc@{}}
\toprule
\multirow{2}{*}{Model}
& \multicolumn{2}{c}{$\tau=0.03$}
& \multicolumn{2}{c}{$\tau=0.05$}
& \multicolumn{2}{c}{$\tau=0.07$} \\
\cmidrule(lr){2-3}
\cmidrule(lr){4-5}
\cmidrule(lr){6-7}
& IF & NL & IF & NL & IF & NL \\
\midrule

T2I
& 59.40 & 56.70
& 66.65  & \textbf{78.10}
& 96.10 & 82.80 \\

T2I-VIF
& \textbf{83.95} & \textbf{61.60}
& \textbf{92.20} & 70.00
& \textbf{100.00} & \textbf{83.90} \\

MLP
& 69.10 & 54.20
& 78.55 & 68.75
& \textbf{100.00} & 69.65 \\

TabTransformer
& 63.80 & 54.70
& 79.50 & 62.50
& 98.50 & 64.55 \\

XGBoost
& 18.90 & 48.45 
& 33.35 & 58.35
& 66.65 & 59.35 \\

SVM
& 10.00 & 52.10
& 25.80 & 56.25
& 66.65 & 62.50 \\

\midrule
\multicolumn{7}{c}{SD} \\
\cmidrule(lr){1-7}

Model
& \multicolumn{2}{c}{$\tau=0.03$}
& \multicolumn{2}{c}{$\tau=0.05$}
& \multicolumn{2}{c}{$\tau=0.07$} \\
\midrule

T2I
& \multicolumn{2}{c}{$(87.67,\mathbf{65.00})$}
& \multicolumn{2}{c}{$(90.00,76.92)$}
& \multicolumn{2}{c}{$(92.73,77.78)$} \\

T2I-VIF
& \multicolumn{2}{c}{$(\mathbf{88.00},63.33)$}
& \multicolumn{2}{c}{$(90.00,\mathbf{79.17})$}
& \multicolumn{2}{c}{$(\mathbf{93.64},\mathbf{90.00})$} \\

MLP
& \multicolumn{2}{c}{$(85.77,47.12)$}
& \multicolumn{2}{c}{$(\mathbf{91.15},60.00)$}
& \multicolumn{2}{c}{$(\mathbf{93.64},62.50)$} \\

TabTransformer
& \multicolumn{2}{c}{$(83.33,36.66)$}
& \multicolumn{2}{c}{$(89.00,40.83)$}
& \multicolumn{2}{c}{$(91.43,61.54)$} \\

SVM
& \multicolumn{2}{c}{$(79.00,34.38)$}
& \multicolumn{2}{c}{$(83.93,39.77)$}
& \multicolumn{2}{c}{$(86.25,60.83)$} \\

XGBoost
& \multicolumn{2}{c}{$(80.31,13.28)$}
& \multicolumn{2}{c}{$(85.63,16.66)$}
& \multicolumn{2}{c}{$(88.44,20.54)$} \\

\bottomrule
\end{tabular}
\end{table}
\else
  
\begin{table}[!hb]
\centering
\caption{Robustness results across three corruption scenarios. We compute the metrics for each dataset and run and report the average. Higher values are better for all metrics, and MLP denotes parameter-matched MLP. 
}
\label{tab:robustness_summary}
\setlength{\tabcolsep}{5pt}
\renewcommand{\arraystretch}{1.10}

\begin{tabular}{@{}llcc@{}}
\toprule
Scenario & Model
& $\operatorname{APS}_{\mathrm{ACC/AUC}}\uparrow$
& $\operatorname{WSP}_{\mathrm{ACC/AUC}}\uparrow$ \\
\midrule

\multirow{6}{*}{IF}
& T2I            & $\mathbf{0.863}/\mathbf{0.929}$ & $0.854/0.924$ \\
& T2I-VIF        & $\mathbf{0.863}/\mathbf{0.929}$ & $\mathbf{0.855}/\mathbf{0.925}$ \\
& MLP            & $0.857/0.927$ & $0.851/0.922$ \\
& TabTransformer & $0.842/0.923$ & $0.840/0.922$ \\
& XGBoost        & $0.839/0.924$ & $0.837/0.921$ \\
& SVM            & $0.828/0.906$ & $0.824/0.904$ \\
\midrule

\multirow{6}{*}{NL}
& T2I            & $\mathbf{0.846}/\mathbf{0.922}$ & $\mathbf{0.822}/\mathbf{0.910}$ \\
& T2I-VIF        & $0.844/0.920$ & $0.815/0.908$ \\
& MLP            & $0.844/0.919$ & $\mathbf{0.822}/0.904$ \\
& XGBoost        & $0.828/0.910$ & $0.807/0.890$ \\
& SVM            & $0.826/0.899$ & $0.817/0.891$ \\
& TabTransformer & $0.824/0.911$ & $0.812/0.902$ \\
\midrule

\multirow{6}{*}{SD}
& T2I            & $\mathbf{0.853}/\mathbf{0.920}$ & $0.721/0.812$ \\
& T2I-VIF        & $\mathbf{0.853}/\mathbf{0.920}$ & $\mathbf{0.722}/\mathbf{0.813}$ \\
& MLP            & $0.849/0.919$ & $0.718/0.804$ \\
& TabTransformer & $0.818/0.905$ & $0.646/0.752$ \\
& SVM            & $0.810/0.890$ & $0.672/0.758$ \\
& XGBoost        & $0.797/0.898$ & $0.460/0.701$ \\

\bottomrule
\end{tabular}
\end{table}

\begin{table}[!t]
\centering
\caption{$\rho_{\tau}^{\mathrm{AUC}}$ (\%) values across three corruption scenarios for $\tau \in \{0.03, 0.05, 0.07\}$. Higher values are better; values are reported as normalized $(\rho_s,a_s)$ percentages for SD. $\rho_{\tau}^{\mathrm{ACC}}$ results are reported in~\cite{t2i_extended}.}
\label{tab:tau_auc}
\setlength{\tabcolsep}{2.3pt}
\renewcommand{\arraystretch}{1.04}

\begin{tabular}{@{}lcccccc@{}}
\toprule
\multirow{2}{*}{Model}
& \multicolumn{2}{c}{$\tau=0.03$}
& \multicolumn{2}{c}{$\tau=0.05$}
& \multicolumn{2}{c}{$\tau=0.07$} \\
\cmidrule(lr){2-3}
\cmidrule(lr){4-5}
\cmidrule(lr){6-7}
& IF & NL & IF & NL & IF & NL \\
\midrule

T2I
& 59.40 & 56.70
& 66.65  & \textbf{78.10}
& 96.10 & 82.80 \\

T2I-VIF
& \textbf{83.95} & \textbf{61.60}
& \textbf{92.20} & 70.00
& \textbf{100.00} & \textbf{83.90} \\

MLP
& 69.10 & 54.20
& 78.55 & 68.75
& \textbf{100.00} & 69.65 \\

TabTransformer
& 63.80 & 54.70
& 79.50 & 62.50
& 98.50 & 64.55 \\

XGBoost
& 18.90 & 48.45 
& 33.35 & 58.35
& 66.65 & 59.35 \\

SVM
& 10.00 & 52.10
& 25.80 & 56.25
& 66.65 & 62.50 \\

\midrule
\multicolumn{7}{c}{SD} \\
\cmidrule(lr){1-7}

Model
& \multicolumn{2}{c}{$\tau=0.03$}
& \multicolumn{2}{c}{$\tau=0.05$}
& \multicolumn{2}{c}{$\tau=0.07$} \\
\midrule

T2I
& \multicolumn{2}{c}{$(87.67,\mathbf{65.00})$}
& \multicolumn{2}{c}{$(90.00,76.92)$}
& \multicolumn{2}{c}{$(92.73,77.78)$} \\

T2I-VIF
& \multicolumn{2}{c}{$(\mathbf{88.00},63.33)$}
& \multicolumn{2}{c}{$(90.00,\mathbf{79.17})$}
& \multicolumn{2}{c}{$(\mathbf{93.64},\mathbf{90.00})$} \\

MLP
& \multicolumn{2}{c}{$(85.77,47.12)$}
& \multicolumn{2}{c}{$(\mathbf{91.15},60.00)$}
& \multicolumn{2}{c}{$(\mathbf{93.64},62.50)$} \\

TabTransformer
& \multicolumn{2}{c}{$(83.33,36.66)$}
& \multicolumn{2}{c}{$(89.00,40.83)$}
& \multicolumn{2}{c}{$(91.43,61.54)$} \\

SVM
& \multicolumn{2}{c}{$(79.00,34.38)$}
& \multicolumn{2}{c}{$(83.93,39.77)$}
& \multicolumn{2}{c}{$(86.25,60.83)$} \\

XGBoost
& \multicolumn{2}{c}{$(80.31,13.28)$}
& \multicolumn{2}{c}{$(85.63,16.66)$}
& \multicolumn{2}{c}{$(88.44,20.54)$} \\

\bottomrule
\end{tabular}
\end{table}
\fi

\begin{figure}[!t]
\begin{center}
\centerline{\includegraphics[width=\linewidth]{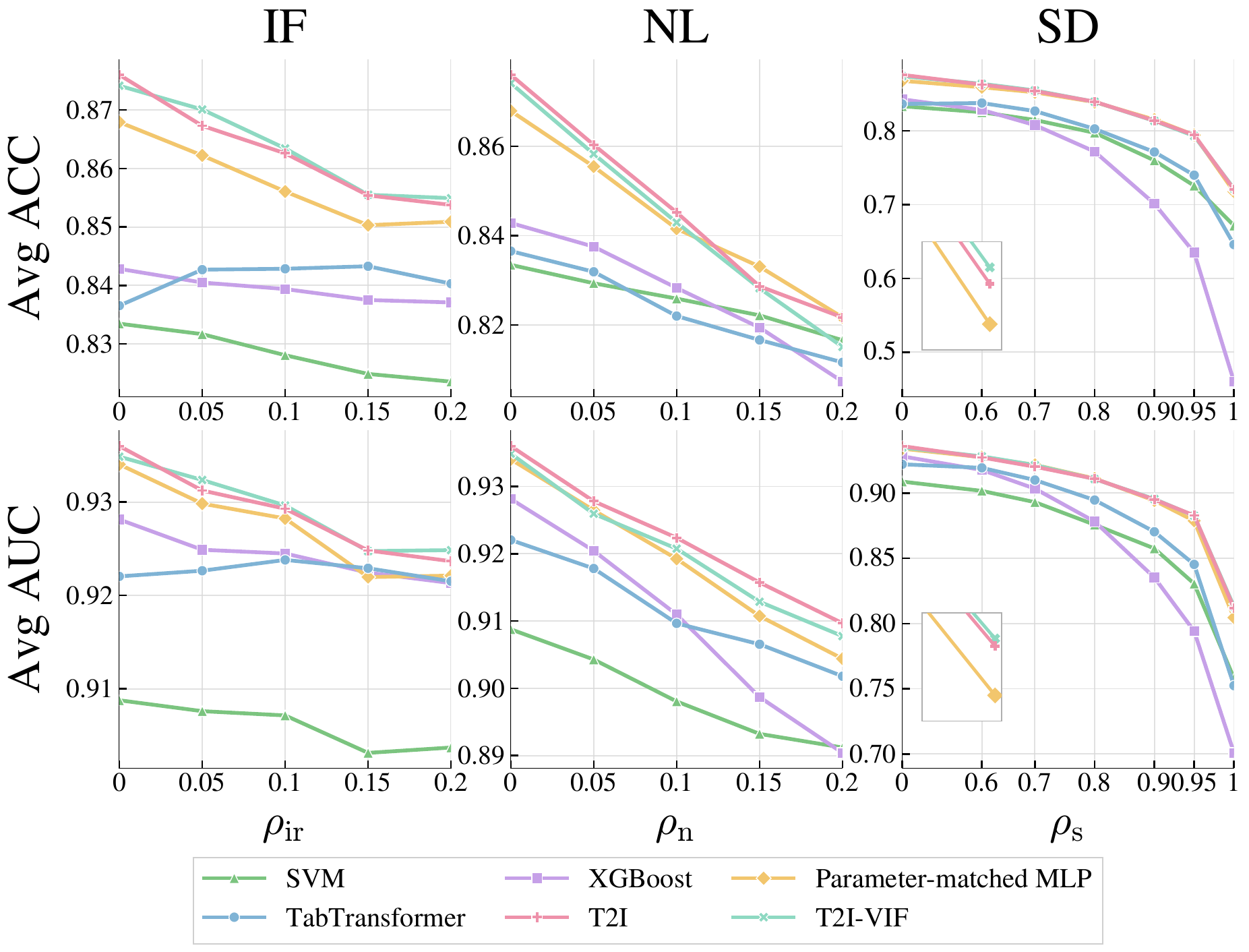}} 
\caption{Average ACC and AUC across increasing corruption severity in three scenarios. For shortcut decorrelation, ACC and AUC are averaged over $a_s$.}
\label{fig:aps_wsp}
\end{center}
\end{figure}

\texttt{T2I} and T2I-VIF provide favorable overall robustness across the three corruption scenarios in accuracy and AUC, relative to tree-based and transformer baselines (Table~\ref{tab:robustness_summary}). In particular, they attain the highest or tied-highest $\text{APS}$ and $\text{WSP}$ values across most settings, showing favorable absolute performance across corruption severity levels and competitive performance under the most severe corruption. The largest observed differences occur under shortcut decorrelation, where both variants outperform every baseline in average $\text{APS}$ and $\text{WSP}$ for both accuracy and AUC (Figure~\ref{fig:aps_wsp}). The $\rho_{\tau}^{\mathrm{AUC}}$ results (Table~\ref{tab:tau_auc}) are mixed across metrics and scenarios, indicating that higher absolute performance may not uniformly correspond to slower relative degradation (see~\extref{Table~\ref{tab:tau_acc}} for $\rho_{\tau}^{\mathrm{ACC}}$). Relative to the parameter-matched MLP, \texttt{T2I} largely preserves its modest clean performance advantage under corruption, although it does not consistently degrade more slowly in relative terms; its higher $\rho_{\tau}$ values indicate greater tolerance to specific degradation thresholds rather than uniformly slower decline.

\paragraph{Dataset-level paired clean-versus-corrupted comparison}

\begin{figure}[!t]
\begin{center}
\centerline{\includegraphics[width=\linewidth]{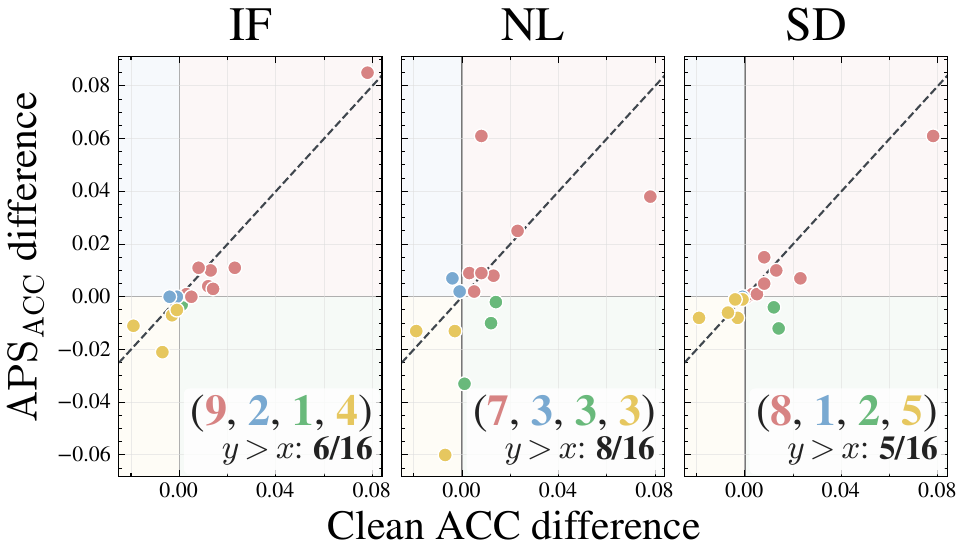}} 
\caption{Dataset-level comparison of \texttt{T2I} and the parameter-matched MLP across scenarios, with each point representing one dataset. Axes show \texttt{T2I}$-$MLP differences in clean accuracy and $\operatorname{APS}_{\mathrm{ACC}}$; the diagonal indicates whether the clean advantage is preserved under corruption.}
\label{fig:clean_vs_mlp}
\end{center}
\end{figure}

Figure~\ref{fig:clean_vs_mlp} compares \texttt{T2I} with the parameter-matched MLP at the dataset level, plotting differences in clean accuracy ($x$-axis) against $\operatorname{APS}_{\mathrm{ACC}}$ ($y$-axis). The upper-right (red) quadrant marks datasets where \texttt{T2I} has higher performance than the parameter-matched MLP in both clean accuracy and $\operatorname{APS}_{\mathrm{ACC}}$: 9, 7, and 8 of 16 datasets for IF, NL, and SD, respectively. Including the upper-left (blue) quadrant, where \texttt{T2I} underperforms on clean accuracy but leads on $\operatorname{APS}_{\mathrm{ACC}}$, this rises to 11, 10, and 9 datasets with a corruption-integrated advantage. On 6, 8, and 5 datasets ($y>x$), the advantage of \texttt{T2I} over the MLP is larger under corruption than on clean data. Overall, \texttt{T2I} holds a positive corruption-integrated advantage on a majority of datasets in every scenario, though how much this advantage grows relative to its clean data edge varies across datasets and scenarios.

\paragraph{Realization-Level Instability}

Using $K=3$ independent realizations within each of three training seeds, we evaluate within-scenario instability using the metrics proposed in Section~\ref{sec:evaluation_metrics}. \texttt{T2I} achieves the lowest ${\operatorname{SAI}}$ and the second-lowest ${\operatorname{DPI}}$, but shows higher ${\operatorname{LCI}}$ (Table~\ref{tab:integrated_instability_combined}). Classical machine learning models, including SVM and XGBoost, show relatively low observed instability in several settings. Among the evaluated deep models (MLP, TabTransformer, and \texttt{T2I}), \texttt{T2I} shows particularly low ${\operatorname{SAI}}$, while its ${\operatorname{LCI}}$ remains comparatively high. 
These results highlight realization-level differences that are not captured by performance-based robustness metrics alone; in particular, the relatively high ${\operatorname{LCI}}$ of \texttt{T2I} despite favorable predictive performance illustrates the complementary information provided by instability evaluation.

\begin{table}[!b]
\centering
\caption{Severity-integrated instability at $K=3$ using all 16 datasets and its maximum absolute variation across $K\in\{3,5,10\}$ on five representative datasets (OpenML IDs: 11, 31, 1468, 1494, and 4134). For severity-integrated instability, the best results are shown in bold, and the second-best results are underlined. Lower values are better. MLP denotes the parameter-matched MLP.}
\label{tab:integrated_instability_combined}
\setlength{\tabcolsep}{3.2pt}
\renewcommand{\arraystretch}{1.08}

\begin{tabular}{lcccccc}
\toprule
\multirow{2}{*}{Model}
& \multicolumn{2}{c}{${\operatorname{DPI}}\downarrow$}
& \multicolumn{2}{c}{${\operatorname{LCI}}\downarrow$}
& \multicolumn{2}{c}{${\operatorname{SAI}}\downarrow$} \\
\cmidrule(lr){2-3}
\cmidrule(lr){4-5}
\cmidrule(lr){6-7}
& $K{=}3$ & $\Delta_{\max}$
& $K{=}3$ & $\Delta_{\max}$
& $K{=}3$ & $\Delta_{\max}$ \\
\midrule
SVM
& \textbf{0.081} & 0.006
& \textbf{0.163} & 0.003
& 0.288 & 0.001 \\
XGBoost
& \textbf{0.081} & 0.001
& \underline{0.182} & 0.004
& 0.309 & 0.005 \\
MLP
& 0.127 & 0.003
& 0.257 & 0.005
& \underline{0.197} & 0.003 \\
TabTransformer
& 0.138 & 0.021
& 0.236 & 0.045
& 0.267 & 0.002 \\
{T2I}
& \underline{0.114} & 0.003
& 0.291 & 0.002
& \textbf{0.190} & 0.003 \\
\bottomrule
\end{tabular}
\end{table}

\textbf{Sensitivity to the number of realizations.}
For five representative datasets listed in Table~\ref{tab:integrated_instability_combined}, selected to cover different sample sizes, dimensionalities, class counts, and observed instability levels, we assess sensitivity across $K\in\{3,5,10\}$ using
$\Delta_{\max}=\max_{K,K'}|I_K-I_{K'}|$, the maximum absolute difference across the evaluated values of $K$, where $I_K$ is the instability obtained using $K$ realizations. Instability is computed within each training seed and then averaged across seeds, separating realization-level variability from training stochasticity. The results remain generally stable, and model rankings are identical across $K$ for ${\operatorname{DPI}}$ and ${\operatorname{SAI}}$; for ${\operatorname{LCI}}$, only the ordering of \texttt{T2I} and TabTransformer changes. The largest variation occurs for ${\operatorname{LCI}}$ of TabTransformer, while for \texttt{T2I}, $\Delta_{\max}$ is at most $0.003$ across all three metrics. These results support $K=3$ as a reasonable approximation for the full-dataset analysis; details are in~\extref{Table~\ref{tab:instability_scaling}}.

\paragraph{Robustness is not due to underfitting}
The robustness gains are unlikely to arise simply from weaker fitting of the corrupted training data. Across all scenarios, \texttt{T2I} (internal proxy targets) achieves the highest training performance among the compared models, with training ACC/AUC exceeding $0.99$ under irrelevant features and shortcut decorrelation, and $0.92/0.96$ under noisy labels. T2I-VIF consistently ranks second with similarly high training performance.
The favorable test-time performance of \texttt{T2I} is thus not accompanied by weaker fitting of the corrupted training data in these experiments.  See~\extref{Table~\ref{tab:training_fit_robustness}} for details.

\subsubsection{Design Analysis: drivers of \texttt{T2I}}
\label{sec:results:ablation}

\paragraph{Why not a single mapping?} \label{sec:results:single}
HACNet \cite{matsuda2024hacnet} assigns each class a single representative image; \texttt{T2I} (external proxy targets) instead employs randomized mappings from diverse FashionMNIST and MNIST samples. The payoff is visible on OpenML-CC18 datasets with three target classes, where average accuracy and AUC improve from 0.848 and 0.931 (single) to 0.888 and 0.951 (multiple)---evidence that the diversity from the multiple-mapping design contributes. See~\extref{Table~\ref{tab:single_multiple}} for details.

\paragraph{Why VIF Initialization?} \label{sec:experiments:vif}
To justify our specific initialization scheme, we compare the VIF initialization (external proxy targets) with two natural alternatives. T2I-DIR skips concatenation (\ie $\mathtt{P}(x) \oplus \mathtt{P}_{\mathrm{VIF}}(x)$) and uses only $\mathtt{P}_{\mathrm{VIF}}(x)$. T2I-MUL learns a scaling term: $\mathtt{Q}_{\mathrm{VIF}}(x) = c_{\mathrm{mul}} \odot x$ (element-wise multiplication), where $(c_{\mathrm{mul}})_i$ is initialized as $1/{(\mathtt{VIF}_i)}$ and subsequently updated as a trainable parameter.
It is then concatenated as $\mathtt{P}(x) \oplus \mathtt{Q}_{\mathrm{VIF}}(x)$.
T2I-VIF, which combines raw and VIF-processed embeddings, achieves the highest average accuracy on the full OpenML-CC18 datasets among the evaluated variants (0.879), compared with 0.871 for T2I-MUL and 0.609 for T2I-DIR (see~\extref{Table~\ref{tab:vif_initialization}}). These patterns indicate that preserving the original representation is associated with stronger predictive performance, whereas replacing it entirely with a VIF-adjusted representation results in a substantial degradation. This is also consistent with the broader view that correlated variables may still contain useful information and need not be discarded outright.

\paragraph{Learned Generation Pathway Ablations}

To isolate which components of \texttt{T2I} contribute to its performance and robustness, we run three complementary ablations targeting different pieces of the pipeline.

\textbf{Proxy target permutation.}
Given an internally constructed proxy target $q(x,y)$, we sample a random permutation $\pi$ and define the permuted proxy target as
$q_{\pi}(x,y) = \Pi_{\pi}q(x,y),$ where $\Pi_{\pi}$ denotes the corresponding permutation matrix. It aims to disrupt the designed proxy structure, while the same permutation is applied to all classes and samples throughout training and evaluation.
\textbf{Removing the reconstruction loss.}
We remove the proxy reconstruction loss ($\mathcal{L}_{\mathrm{recon}}$) while retaining all other structures. This evaluates the contribution of explicitly aligning the generated representation with the structured proxy target.
\textbf{Generator-free mapping.}
We bypass the generation pathway altogether by constructing a direct $28\times28$ CNN input by repeatedly concatenating the tabular representation, truncating it to 784 dimensions, and applying an element-wise sigmoid.

The generator-free mapping produces the largest and most consistent degradation among the three ablations (Table~\ref{tab:internal_ablation}), supporting the contribution of the learned generation pathway. The proxy target permutation and reconstruction loss removal ablations have only modest effects on clean performance and most robustness metrics.

\begin{table}[!t]
\centering
\caption{Component-wise ablation results for \texttt{T2I} with internally constructed proxy targets, averaged across 16 datasets.}
\label{tab:internal_ablation}
\setlength{\tabcolsep}{2.7pt}
\renewcommand{\arraystretch}{1.08}

\begin{tabular}{@{}llcccc@{}}
\toprule
Scenario & Metric
& Full
& Permuted
& w/o $\mathcal{L}_{\mathrm{recon}}$
& Generator-free \\
\midrule

\multirow{2}{*}{Clean}
& ACC
& \textbf{0.8760}
& 0.8747
& 0.8740
& 0.8607 \\
& AUC
& \textbf{0.9360}
& 0.9350
& 0.9355
& 0.9323 \\

\midrule

\multirow{2}{*}{IF}
& $\operatorname{APS}_{\mathrm{ACC}}$
& \textbf{0.8626}
& \textbf{0.8626}
& \textbf{0.8626}
& 0.8495 \\
& $\operatorname{APS}_{\mathrm{AUC}}$
& \textbf{0.9288}
& 0.9279
& 0.9279
& 0.9256 \\

\midrule

\multirow{2}{*}{NL}
& $\operatorname{APS}_{\mathrm{ACC}}$
& \textbf{0.8458}
& 0.8440
& 0.8439
& 0.8427 \\
& $\operatorname{APS}_{\mathrm{AUC}}$
& \textbf{0.9222}
& 0.9197
& 0.9199
& 0.9216 \\

\midrule

\multirow{2}{*}{SD}
& $\operatorname{APS}_{\mathrm{ACC}}$
& \textbf{0.8529}
& 0.8523
& 0.8516
& 0.8438 \\
& $\operatorname{APS}_{\mathrm{AUC}}$
& \textbf{0.9200}
& 0.9192
& 0.9194
& 0.9180 \\

\bottomrule
\end{tabular}
\end{table}

\paragraph{Does \texttt{T2I} benefit across different encoders?} 

We investigate whether the \texttt{T2I} pathway can serve as a wrapper across different tabular encoders by replacing the original $\mathtt{P}(x)$ while keeping the subsequent architecture unchanged. 
For the MLP-based encoders, we compare a parameter-matched MLP and reduced-capacity variants at 25\% and 50\% of its parameter budget, all retaining the same three-layer structure. 
Across all MLP capacities, the \texttt{T2I} pathway (internal proxy targets) preserves AUC while modestly improving accuracy, with slightly larger gains for the reduced-capacity (25\% and 50\%) variants.
In contrast, applying the \texttt{T2I} pathway to FT-Transformer decreases both ACC and AUC (Table~\ref{tab:t2i_pair_comparison}), although we do not tune the \texttt{T2I} pathway specifically for the transformer encoder. These results suggest that the \texttt{T2I} pathway is more compatible with the evaluated MLP encoders, complementing lightweight MLPs with small accuracy improvements while preserving AUC.

\begin{table}[!b]
\centering
\caption{Clean performance with and without the \texttt{T2I} pathway across MLP- and Transformer-based encoders. MLP capacity is reported relative to the parameter budget of the parameter-matched MLP.}
\label{tab:t2i_pair_comparison}
\setlength{\tabcolsep}{4.0pt}
\renewcommand{\arraystretch}{1.10}

\begin{tabular}{lcccc}
\toprule
\multirow{2}{*}{Encoder}
& \multicolumn{2}{c}{ACC}
& \multicolumn{2}{c}{AUC} \\
\cmidrule(lr){2-3}
\cmidrule(lr){4-5}
& Baseline & {T2I}
& Baseline & {T2I} \\
\midrule
MLP (100\%)
& 0.868 & 0.869
& 0.934 & 0.934 \\
MLP (50\%)
& 0.868 & 0.872
& 0.934 & 0.934 \\
MLP (25\%)
& 0.867 & 0.871
& 0.935 & 0.935 \\
FT-Transformer
& 0.865 & 0.788
& 0.934 & 0.895 \\
\bottomrule
\end{tabular}
\end{table}

\subsection{Analysis of Learned Proxy Representations} \label{sec:experiments:vis}

\begin{figure}[!t]
\begin{center}
\centerline{\includegraphics[width=\linewidth]{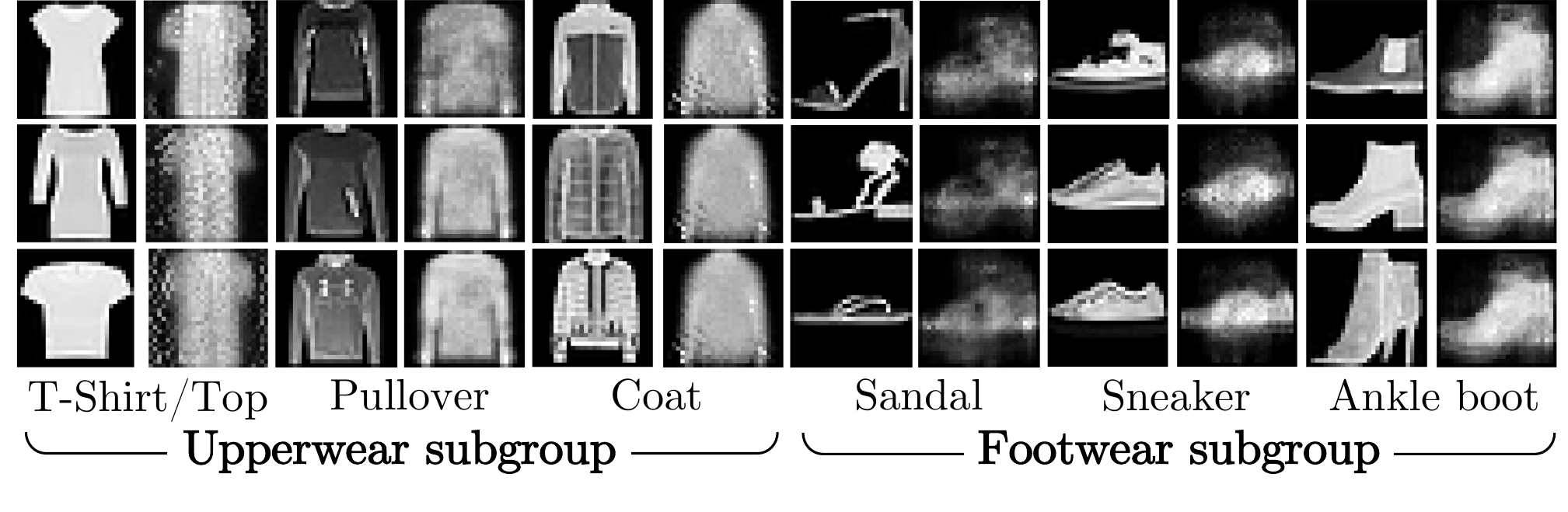}} 
\caption{Examples of generated representations using external proxy targets for correctly classified samples from visually similar upperwear and footwear class subgroups.}
\label{fig:external}
\end{center}
\end{figure}

\begin{figure}[!ht]
\begin{center}
\centerline{\includegraphics[width=\linewidth]{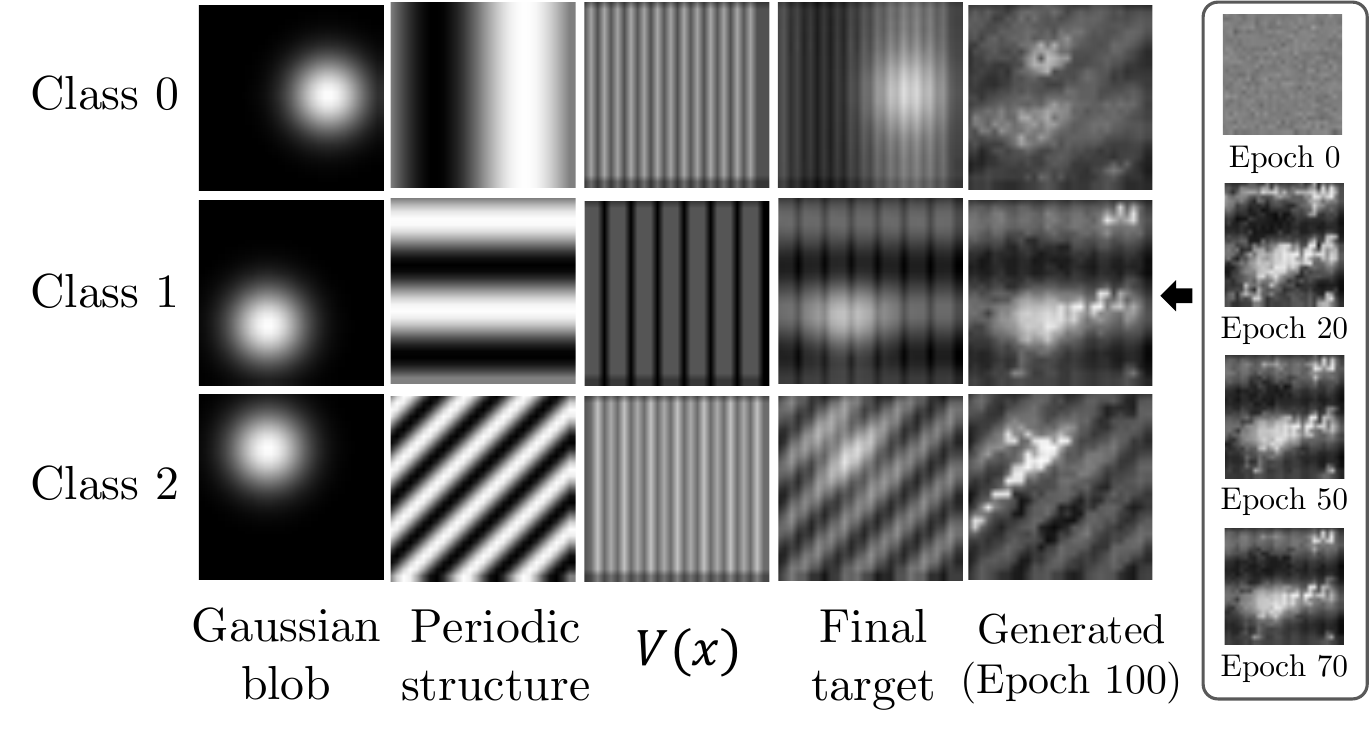}} 
\caption{Examples of internal proxy target construction and the representative generation process for OpenML-CC18 dataset ID 11 with three target classes. The generation example corresponds to Class 1.}
\label{fig:internal}
\end{center}
\end{figure}

We visualize what the learned generation pathway actually produces. Figure~\ref{fig:external} shows generated representations for correctly classified samples under external proxy targets, focusing on class subgroups that are visually difficult to distinguish. Even among closely related categories---(a) T-shirt/top, pullover, and coat, or (b) sandal, sneaker, and ankle boot---the model still produces visually class-distinct patterns.
Under internal proxy targets (Figure~\ref{fig:internal}), the generated representations reflect the class-specific Gaussian and periodic structures while retaining instance-dependent variations from $V(x)$.
Starting from random noise, the generated representation progressively resembles this structured target as training proceeds. 
Some residual artifacts remain, as expected when generating from random noise; however, the objective is not photorealistic reconstruction but the formation of informative proxy representations. The outputs vary across individual inputs while exhibiting a consistent tendency toward their corresponding class-level patterns.

\section{Conclusion}

We introduce \TabletoImage, a lightweight tabular learning model that learns structured proxy representations through a learned generation pathway, together with a VIF-informed initialization variant. Across OpenML-CC18 and TabZilla, \TabletoImage achieves competitive clean predictive performance with fewer parameters than large-scale neural baselines. Under three distinct failure modes---irrelevant features, noisy labels, and shortcut decorrelation---it also maintains favorable performance across corruption severities, although relative degradation and realization-level stability vary across scenarios. In particular, the proposed instability diagnostics reveal behavioral differences not captured by aggregate performance alone, supporting broader reliability evaluation.
Our design analyses further support structured proxy generation as a useful inductive bias for compact tabular learning. 

\textbf{Limitations.} The current study is limited to classification tasks and controlled synthetic corruption scenarios. Additionally, the external proxy target mode supports at most 20 classes. Future work will extend the approach to broader tasks, naturally occurring shifts, and more flexible proxy representations.

\section*{Acknowledgments}
\label{sec:Acknowledgments}

This work was supported by the Korea Institute for Advancement of Technology (KIAT) grant funded by the Korean government (MOTIE) (No.~2410019447, Enterprise Demand-Based 1\% MVP Project), by the National Research Foundation of Korea (NRF) grant funded by the Ministry of Science and ICT (RS-2023-00208284), in part by NSF Awards 2312930 and 2326193, and by NYU IT High Performance Computing resources, services, and staff expertise.

\bibliographystyle{IEEEtran}
\ifarxiv
  \bibliography{refs}
\else
  \bibliography{refs_condensed}
\fi

\ifarxiv
\newpage
\section*{Appendix}

\textbf{Severity-Integrated Instability.} 
For a single severity-dependent instability metric
$
\mathcal{I}(\rho)
\in
\left\{
\operatorname{DPI}(\rho_{\mathrm{ir}}),
\operatorname{LCI}(\rho_n)
\right\},
$
we define its normalized severity-integrated value as
$$
\frac{1}{\rho_{\max}-\rho_{\min}}
\int_{\rho_{\min}}^{\rho_{\max}}
\mathcal{I}(\rho)
\,d\rho.
$$
Accordingly, ${\operatorname{DPI}}$ and
${\operatorname{LCI}}$ summarize instability across $\rho_\text{r}$ and $\rho_\text{n}$, respectively.
For shortcut decorrelation, which varies over both $\rho_\text{s}$ and $a_\text{s}$, we define
$$
\frac{
\displaystyle
\int_{a_\text{s}^{\min}}^{a_\text{s}^{\max}}
\int_{\rho_\text{s}^{\min}}^{\rho_\text{s}^{\max}}
\operatorname{SAI}(\rho_\text{s},a_\text{s})
\,d\rho_\text{s}\,da_\text{s}
}{
\left(\rho_\text{s}^{\max}-\rho_\text{s}^{\min}\right)
\left(a_\text{s}^{\max}-a_\text{s}^{\min}\right)
}.
$$

\textbf{Additional results.}
Figure~\ref{fig:aps_wsp_shortcut_scale} provides shortcut decorrelation results across individual availability levels.
Table~\ref{tab:model_parameters} and Figures~\ref{fig:x_param_y_perf_openml} and~\ref{fig:x_param_y_perf_tabzilla} report detailed parameter counts and parameter-performance comparisons on OpenML-CC18 and TabZilla.
Table~\ref{tab:tabular_to_image_comparison} provides the detailed comparison with existing tabular-to-image methods.

\textbf{Details on T2I-DIR.}
T2I-DIR in Section~\ref{sec:experiments:vif} initializes the first layer weights directly as
$w_{ij} = 1/{(\mathtt{VIF}_i + c_{\mathrm{dir}})}$
with $c_{\mathrm{dir}}=10$ to stabilize small VIF values.

\begin{table}[!ht]
\centering
\caption{Holm-corrected pairwise Wilcoxon signed-rank tests and median paired differences with 95\% confidence intervals for T2I and T2I-VIF against principal baselines. Based on the performance, for TabZilla--AUC, we compare T2I-VIF against the baselines; all other comparisons use T2I.}
\label{tab:pairwise_stats}
\setlength{\tabcolsep}{1.5pt}
\renewcommand{\arraystretch}{1.02}

\begin{tabular}{llccc}
\toprule
Benchmark & Metric & Baseline & $p_{\mathrm{Holm}}$ & Median $\Delta$ (95\% CI) \\
\midrule

\multirow{6}{*}{OpenML-CC18}
& \multirow{3}{*}{ACC}
& TuneTables & 0.091 & 0.018 ([-0.009, 0.024]) \\
& & CatBoost  & 0.073 & 0.012 ([-0.003, 0.018]) \\
& & LightGBM  & 0.046 & 0.019 ([0.001, 0.022]) \\
\cmidrule(lr){2-5}
& \multirow{3}{*}{AUC}
& TuneTables & 0.083 & 0.006 ([-0.011, 0.010]) \\
& & CatBoost  & 0.092 & 0.007 ([-0.006, 0.009]) \\
& & LightGBM  & 0.077 & 0.011 ([-0.002, 0.014]) \\

\midrule

\multirow{6}{*}{TabZilla}
& \multirow{3}{*}{ACC}
& TuneTables & 0.088 & 0.010 ([-0.002, 0.013]) \\
& & CatBoost  & 0.061 & 0.021 ([-0.001, 0.023]) \\
& & LightGBM  & 0.046 & 0.022 ([0.002, 0.025]) \\
\cmidrule(lr){2-5}
& \multirow{3}{*}{AUC}
& TuneTables & 0.068 & 0.009 ([-0.004, 0.017]) \\
& & CatBoost  & 0.072 & 0.005 ([-0.001, 0.007]) \\
& & LightGBM  & 0.049 & 0.004 ([0.002, 0.007]) \\

\bottomrule
\end{tabular}
\end{table}

\begin{figure}[!ht]
\begin{center}
\centerline{\includegraphics[width=\linewidth]{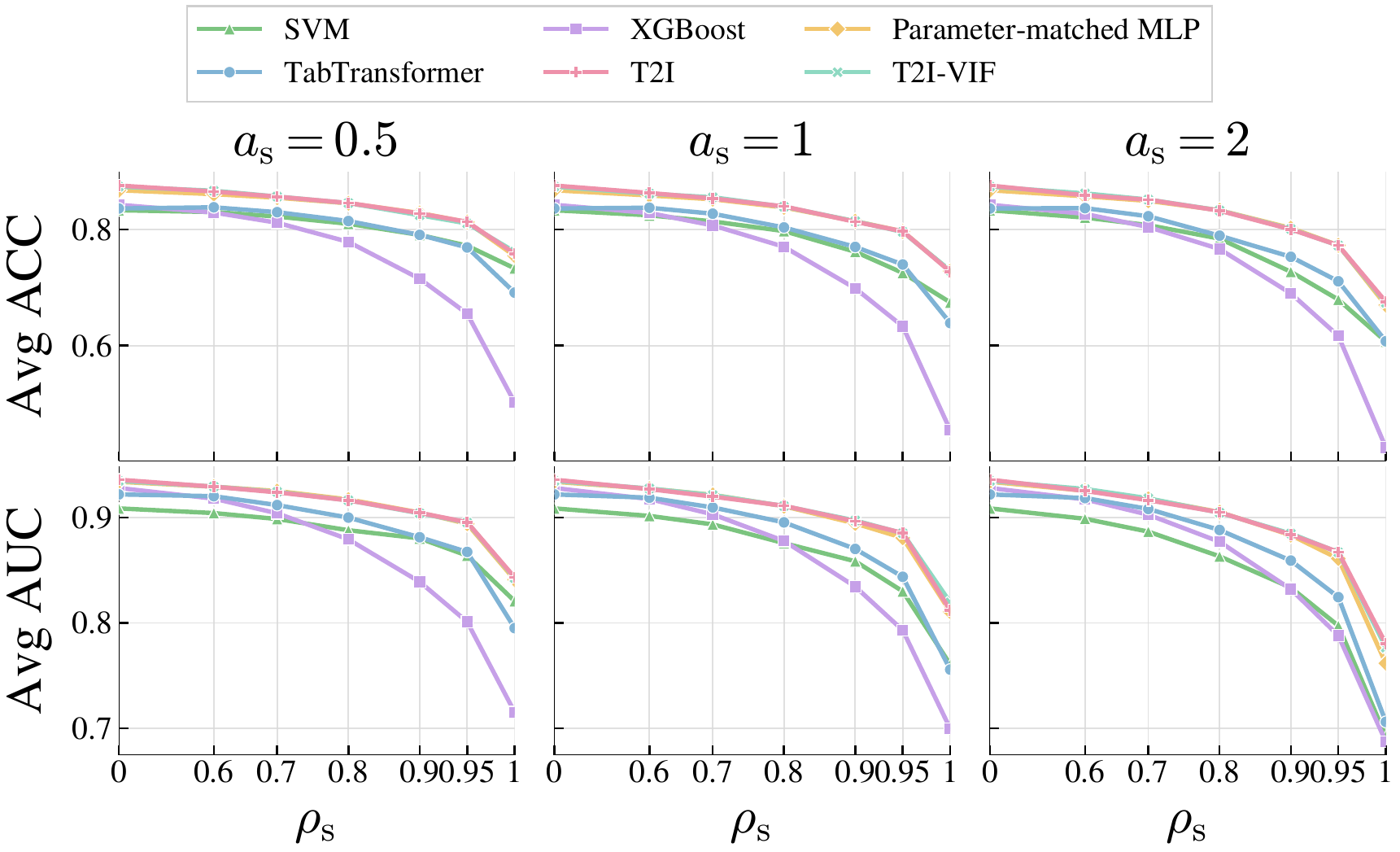}} 
\caption{Average ACC and AUC across increasing shortcut predictivity $\rho_s$ across different availability levels $a_s$, under shortcut decorrelation scenario.}
\label{fig:aps_wsp_shortcut_scale}
\end{center}
\end{figure}

\begin{table}[!ht]
\centering
\caption{Number of trainable parameters for deep learning models with two classes and a tabular input dimension of $N=78$.}
\label{tab:model_parameters}
\setlength{\tabcolsep}{8pt}
\renewcommand{\arraystretch}{1.15}
\begin{tabular}{lc}
\toprule
Model & \# of Parameters \\
\midrule
MLP & 0.007M \\
MLP-PLR & 0.009M \\
FT-Transformer & 0.136M \\
\textbf{Table2Image} & \textbf{0.628M} \\
TabPFN (v1) & 25.8M \\
TuneTables & 25.8M \\
TabM & 37.9M \\
\bottomrule
\end{tabular}
\end{table}

\begin{figure}[!ht]
\begin{center}
\centerline{\includegraphics[width=\linewidth]{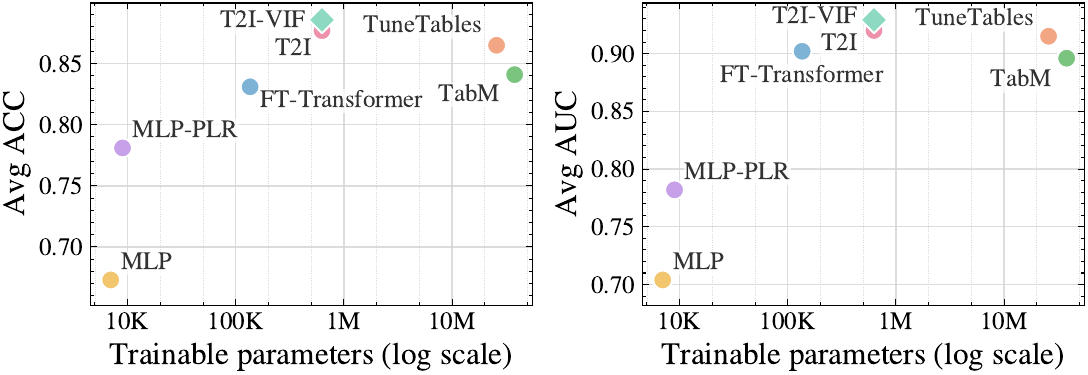}} 
\caption{Average ACC and AUC on OpenML-CC18 datasets versus the number of trainable parameters, with parameter counts shown on a logarithmic scale.}
\label{fig:x_param_y_perf_openml}
\end{center}
\end{figure}

\begin{figure}[!ht]
\begin{center}
\centerline{\includegraphics[width=\linewidth]{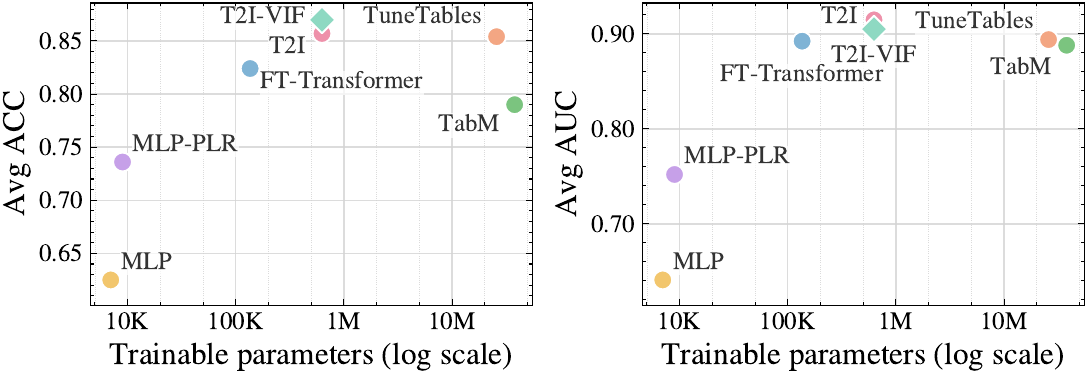}} 
\caption{Average ACC and AUC on TabZilla datasets versus the number of trainable parameters, with parameter counts shown on a logarithmic scale.}
\label{fig:x_param_y_perf_tabzilla}
\end{center}
\end{figure}

\begin{table}[!t]
\centering
\caption{$\rho_{\tau}^{\mathrm{ACC}}$ (\%) values across three corruption scenarios for $\tau \in \{0.03, 0.05, 0.07\}$. Higher values are better; values are reported as normalized $(\rho_s,a_s)$ percentages for SD.}
\label{tab:tau_acc}
\setlength{\tabcolsep}{2.3pt}
\renewcommand{\arraystretch}{1.04}

\begin{tabular}{@{}lcccccc@{}}
\toprule
\multirow{2}{*}{Model}
& \multicolumn{2}{c}{$\tau=0.03$}
& \multicolumn{2}{c}{$\tau=0.05$}
& \multicolumn{2}{c}{$\tau=0.07$} \\
\cmidrule(lr){2-3}
\cmidrule(lr){4-5}
\cmidrule(lr){6-7}
& IF & NL & IF & NL & IF & NL \\
\midrule

T2I
& 38.00 & 39.20
& 69.50 & 51.90
& \textbf{97.65} & 61.55 \\

T2I-VIF
& \textbf{56.65} & 39.20
& \textbf{89.45} & 50.00
& 97.40 & 59.35 \\

MLP
& 55.95 & 44.15
& 85.45 & 58.90
& 90.55 & 66.65 \\

TabTransformer
& 31.55 & 56.25
& 50.00 & 57.30
& 66.65 & 64.55 \\

XGBoost
& 47.50 & 57.50
& 50.00 & 66.35
& 66.65 & \textbf{76.05} \\

SVM
& 6.65 & \textbf{57.85}
& 50.00 & \textbf{67.50}
& 78.15 & 68.75 \\

\midrule
\multicolumn{7}{c}{SD} \\
\cmidrule(lr){1-7}

Model
& \multicolumn{2}{c}{$\tau=0.03$}
& \multicolumn{2}{c}{$\tau=0.05$}
& \multicolumn{2}{c}{$\tau=0.07$} \\
\midrule

T2I
& \multicolumn{2}{c}{$(83.00,37.50)$}
& \multicolumn{2}{c}{$(86.54,56.73)$}
& \multicolumn{2}{c}{$(89.58,77.08)$} \\

T2I-VIF
& \multicolumn{2}{c}{$(83.33,39.28)$}
& \multicolumn{2}{c}{$(87.86,\mathbf{65.18})$}
& \multicolumn{2}{c}{$(90.83,\mathbf{85.42})$} \\

MLP
& \multicolumn{2}{c}{$(\mathbf{84.67},\mathbf{55.00})$}
& \multicolumn{2}{c}{$(\mathbf{90.00},55.77)$}
& \multicolumn{2}{c}{$(\mathbf{92.31},68.75)$} \\

TabTransformer
& \multicolumn{2}{c}{$(79.00,25.90)$}
& \multicolumn{2}{c}{$(83.21,31.66)$}
& \multicolumn{2}{c}{$(87.50,54.81)$} \\

SVM
& \multicolumn{2}{c}{$(76.33,47.65)$}
& \multicolumn{2}{c}{$(81.67,{55.83})$}
& \multicolumn{2}{c}{$(85.36,58.93)$} \\

XGBoost
& \multicolumn{2}{c}{$(76.25,12.50)$}
& \multicolumn{2}{c}{$(80.31,12.50)$}
& \multicolumn{2}{c}{$(85.31,13.33)$} \\

\bottomrule
\end{tabular}
\end{table}

\begin{table}[!t]
\centering
\caption{Severity-integrated instability under different numbers of realizations. MLP denotes the parameter-matched MLP.}
\label{tab:instability_scaling}
\setlength{\tabcolsep}{3.5pt}
\renewcommand{\arraystretch}{1.10}

\begin{tabular}{llccc}
\toprule
Model & Level
& ${\operatorname{DPI}}\downarrow$
& ${\operatorname{LCI}}\downarrow$
& ${\operatorname{SAI}}\downarrow$ \\
\midrule

\multirow{3}{*}{SVM}
& $K=3$  & 0.0854 & 0.1448 & 0.2460 \\
& $K=5$ & 0.0883 & 0.1421 & 0.2451 \\
& $K=10$  & 0.0909 & 0.1416 & 0.2465 \\
\midrule

\multirow{3}{*}{XGBoost}
& $K=3$  & 0.1206 & 0.2323 & 0.3279 \\
& $K=5$ & 0.1201 & 0.2341 & 0.3273 \\
& $K=10$  & 0.1197 & 0.2361 & 0.3234 \\
\midrule

\multirow{3}{*}{MLP}
& $K=3$  & 0.1275 & 0.2758 & 0.1880 \\
& $K=5$ & 0.1251 & 0.2779 & 0.1862 \\
& $K=10$  & 0.1250 & 0.2804 & 0.1850 \\
\midrule

\multirow{3}{*}{TabTransformer}
& $K=3$  & 0.1552 & 0.2806 & 0.2268 \\
& $K=5$ & 0.1764 & 0.3101 & 0.2244 \\
& $K=10$  & 0.1762 & 0.3256 & 0.2262 \\
\midrule

\multirow{3}{*}{\texttt{T2I}}
& $K=3$  & 0.1131 & 0.2816 & 0.1533 \\
& $K=5$ & 0.1155 & 0.2815 & 0.1527 \\
& $K=10$  & 0.1161 & 0.2839 & 0.1506 \\

\bottomrule
\end{tabular}
\end{table}

\begin{table}[!ht]
\caption{Performance comparison between multiple random image mapping and single image mapping using Table2Image on datasets from the OpenML-CC18 benchmark suite with three target classes.}
\label{tab:single_multiple}
\vskip 0.15in
\begin{center}
    \renewcommand{\arraystretch}{1.2} 
    \setlength{\tabcolsep}{1.8pt}  
    \begin{tabular}{c*{2}{cc}*{1}{c}}
        \toprule
         & \multicolumn{2}{c}{Multiple} & \multicolumn{2}{c}{Single} \\
         \cmidrule(lr){2-3} \cmidrule(lr){4-5}
         Dataset & \ ACC \ & AUC & \ ACC \ & AUC \\
        \midrule
        \ balance-scale \ & 1.0000 & \ 1.0000 \ & 0.9520 & \ 0.9886 \ \\
        cmc & 0.6000 & 0.7622 & 0.5932 & 0.7635 \\
        splice & 0.9577 & 0.9894 & 0.8840 & 0.9563 \\
        connect-4 & 0.8336 & 0.9578 & 0.8153 & 0.8982 \\
        jungle-chess & 0.9656 & 0.9978 & 0.8843 & 0.9815 \\
        dna & 0.9687 & 0.9962 & 0.9608 & 0.9951 \\
        \midrule
        Average & \textbf{0.8876} & \textbf{0.9506} & 0.8483 & 0.9305 \\
        \# of Wins & \textbf{6} & \textbf{5} & 0 & 1 \\
        \bottomrule
    \end{tabular}
    \end{center}
\end{table}

\begin{table}[!ht]
\centering
\caption{Performance comparison of Table2Image (T2I) and Table2Image-VIF (T2I-VIF) with existing tabular-to-image methods.}
\label{tab:tabular_to_image_comparison}
\renewcommand{\arraystretch}{1.15}
\setlength{\tabcolsep}{3pt}

\textbf{(a) Comparison with DeepInsight and HACNet}

\vspace{0.05in}

\begin{tabular}{lccc}
\toprule
Dataset & T2I & DeepInsight & HACNet \\
\midrule
balance-scale & \textbf{1.0000} & 0.8457 & 0.8571 \\
cmc           & \textbf{0.6000} & 0.5860 & 0.5405 \\
splice        & \textbf{0.9577} & 0.8882 & 0.9091 \\
connect-4     & \textbf{0.8336} & 0.7636 & 0.7134 \\
jungle-chess  & \textbf{0.9656} & 0.7973 & 0.7019 \\
dna           & \textbf{0.9687} & 0.9393 & 0.9624 \\
\midrule
Avg ACC       & \textbf{0.8876} & 0.8034 & 0.7807 \\
\# Wins       & \textbf{6} & 0 & 0 \\
\bottomrule
\end{tabular}

\vspace{0.15in}

\textbf{(b) Comparison with IGTD}

\vspace{0.05in}

\begin{tabular}{lcc}
\toprule
Dataset (\# Classes) & T2I & IGTD \\
\midrule
Internet-Advertisements (2) & 0.9873 & \textbf{0.9970} \\
har (6)                     & \textbf{0.9998} & 0.9985 \\
MiceProtein (8)             & \textbf{1.0000} & 0.9954 \\
cnae (9)                    & \textbf{0.9923} & 0.8935 \\
\midrule
Avg ACC                     & \textbf{0.9949} & 0.9711 \\
\# Wins                     & \textbf{3} & 1 \\
\bottomrule
\end{tabular}

\vspace{0.15in}

\textbf{(c) Aggregated comparison with HACNet}

\vspace{0.05in}

\begin{tabular}{lcc}
\toprule
Model & Avg ACC & \# Wins \\
\midrule
T2I & 0.8766 & 41 \\
T2I-VIF & \textbf{0.8787} & \textbf{53} \\
HACNet & 0.7999 & 5 \\
\bottomrule
\end{tabular}

\end{table}

\begin{table}[!ht]
\centering
\caption{Comparison of T2I, T2I-VIF (under external proxy targets) and VIF initialization variants T2I-DIR and T2I-MUL.}
\label{tab:vif_initialization}
\footnotesize
\renewcommand{\arraystretch}{1.15}
\setlength{\tabcolsep}{5pt}
\begin{tabular}{lcccc}
\toprule
\multirow{2}{*}{Model}
& \multirow{2}{*}{Avg ACC}
& \multirow{2}{*}{Avg AUC}
& \multicolumn{2}{c}{\# Wins} \\
\cmidrule(lr){4-5}
& & & ACC & AUC \\
\midrule
T2I
& 0.877 & 0.920 & 37 & 40 \\
T2I-VIF
& \textbf{0.879} & \textbf{0.922}
& \textbf{44} & \textbf{41} \\
T2I-DIR
& 0.609 & 0.781 & 2 & 8 \\
T2I-MUL
& 0.871 & 0.917 & 33 & 25 \\
\bottomrule
\end{tabular}
\end{table}

\begin{table}[!t]
\centering
\caption{Train accuracy and AUC under robustness scenarios averaged across datasets, with ranks computed among all six compared models.}
\label{tab:training_fit_robustness}
\setlength{\tabcolsep}{3pt}
\renewcommand{\arraystretch}{1.08}
\begin{tabular}{llcc}
\toprule
Scenario & Model & Train ACC (Rank) & Train AUC (Rank) \\
\midrule
\multirow{2}{*}{IF}
& T2I     & \textbf{0.9922 (1/6)} & \textbf{0.9993 (1/6)} \\
& T2I-VIF & 0.9922 (2/6)          & 0.9993 (2/6) \\

\midrule
\multirow{2}{*}{NL}
& T2I     & \textbf{0.9262 (1/6)} & \textbf{0.9687 (1/6)} \\
& T2I-VIF & 0.9223 (2/6)          & 0.9674 (2/6) \\

\midrule
\multirow{2}{*}{SD}
& T2I     & \textbf{0.9907 (1/6)} & \textbf{0.9988 (1/6)} \\
& T2I-VIF & 0.9901 (2/6)          & 0.9986 (2/6) \\
\bottomrule
\end{tabular}
\end{table}

\fi

\end{document}